\newif\ifarxiv 
\arxivtrue

\newif\ifreview 
\reviewfalse

\newif\iffinal 
\finaltrue

\ifarxiv
\documentclass{article}
\else
\documentclass[envcountsame]{llncs}
\fi

\usepackage[utf8]{inputenc}
\usepackage[utf8]{inputenc}
\usepackage{svg}
\ifarxiv
\usepackage[a4paper,left=2cm,right=2cm,top=2cm,bottom=2cm]{geometry}
\usepackage[onehalfspacing]{setspace}
\usepackage{amsthm}
\usepackage{authblk}
\usepackage[hidelinks]{hyperref}
\usepackage[numbers]{natbib}
\fi

\usepackage{amsmath,amssymb,mathtools,dsfont}

\usepackage{booktabs} 
\usepackage{xspace}
\usepackage{graphicx}
\usepackage{xspace}
\usepackage{url}
\usepackage{enumitem}
\setlist[enumerate,1]{label=\textup{(\roman*)}}

\usepackage{caption} 
\usepackage{subcaption}
\usepackage{booktabs}
\usepackage{svg}
\usepackage{multirow}

\usepackage[algo2e,ruled,vlined,linesnumbered]{algorithm2e} 
\SetAlgoSkip{}
\DontPrintSemicolon

\ifarxiv 
\usepackage[appendix=inline]{apxproof}  
\RenewEnviron{proofsketch}{}
\RenewEnviron{toappendix}{}
\else
\ifreview
\usepackage[appendix=append]{apxproof}  
\else
\usepackage[appendix=strip]{apxproof}   
\fi
\usepackage{xparse} 
\RenewDocumentEnvironment{proofsketch}{o} 
{%
  \par\noindent\textit{Proof sketch%
  \IfValueT{#1}{ (#1)}.}\quad
}
{\par}
\fi

\allowdisplaybreaks[3] 

\usepackage{tikz}
\usetikzlibrary{shapes,arrows}
\usetikzlibrary{decorations}
\usetikzlibrary{decorations.markings}
\usetikzlibrary{plotmarks}
\usetikzlibrary{calc}
\usetikzlibrary{mindmap}
\usetikzlibrary{shadows}
\usetikzlibrary{backgrounds}
\usetikzlibrary{patterns}
\usetikzlibrary{shapes.symbols}
\usetikzlibrary{shapes.gates.logic.US}

\usepackage{pgfplots}
\usepackage{pgfplotstable}
\usepgfplotslibrary{statistics}
\pgfplotsset{compat=1.18}

\ifarxiv
\newtheorem{theorem}             {Theorem}
\newtheorem{lemma}      [theorem]{Lemma}

\fi

\newtheoremrep{theorem}{Theorem}
\newtheoremrep{lemma}     [theorem]{Lemma}
\newtheoremrep{definition}[theorem]{Definition}
\newtheoremrep{conjecture}[theorem]{Conjecture}
\newtheoremrep{corollary} [theorem]{Corollary}
\newtheoremrep{proposition}[theorem]{Proposition}

\ifarxiv\else
\AtEndEnvironment{proof}{\qed}
\AtEndEnvironment{proofsketch}{\qed}
\fi

\usepackage{tabularx,colortbl}

\makeatletter
\g@addto@macro\bfseries{\boldmath}
\makeatother

\newcommand{\ie}{i.\,e.\xspace}
\newcommand{\wrt}{w.\,r.\,t.\xspace}
\newcommand{\eg}{e.\,g.\xspace}

\newcommand{\Z}{\mathbb{Z}}
\newcommand{\N}{\mathbb{N}}
\newcommand{\R}{\mathbb{R}}

\DeclareMathOperator{\XOR}{\oplus} 

\DeclareMathOperator{\Unif}{Unif}                         
\DeclareMathOperator{\Geom}{Geom}                         

\newcommand*{\bigO}{\mathcal{O}}

\DeclareMathOperator{\poly}{poly}                         

\newcommand{\prob}[1]{\Pr\left(#1\right)}                 
\newcommand{\expect}[1]{\mathrm{E}\left[#1\right]}        

\newcommand{\termset}{\ensuremath{T}}                            
\newcommand{\funcset}{\ensuremath{F}}                            

\newcommand{\searchspace}{\ensuremath{\mathcal{X}}\xspace}       

\newcommand{\Nin}{\ensuremath{N_{\mathrm{in}}}}                  
\newcommand{\Nout}{\ensuremath{N_{\mathrm{out}}}}                
\newcommand{\Nfunc}{\ensuremath{N_{\mathrm{func}}}}              
\newcommand{\Nall}{\ensuremath{N}}                               
\newcommand{\Kfunc}{\ensuremath{K}}                              
\newcommand{\arity}{\ensuremath{k}}                              

\newcommand{\opotgp}{{\upshape($1$+$1$)~TGP}\xspace}                       
\newcommand{\opotgpstrict}{{\upshape($1$+$1$)~TGP$^*$}\xspace}             

\newcommand{\opocgp}{{\upshape($1$+$1$)~CGP}\xspace}                       
\newcommand{\opocgpstrict}{{\upshape($1$+$1$)~CGP$^*$}\xspace}             

\newcommand{\maxproblem}{$\textsc{Max}$\xspace}                                      

\newcommand{\inputi}[1]{\ensuremath{x_{#1}}\xspace}             
\newcommand{\bool}{\ensuremath{\textsc{Bool}}\xspace}           
\newcommand{\valtrue}{\ensuremath{\textsc{True}}\xspace}        
\newcommand{\valfalse}{\ensuremath{\textsc{False}}\xspace}      
\newcommand{\funcbinand}{\ensuremath{\textsc{And}}\xspace}      
\newcommand{\funcbinxor}{\ensuremath{\textsc{Xor}}\xspace}      

\newcommand{\andproblem}{$\textsc{And}_n$\xspace}               
\newcommand{\andproblemparam}[2]{$\textsc{And}_n$-#1-#2\xspace} 
\newcommand{\andproblemfull}{\andproblemparam{$\{\textsc{And}\}$}{$\{x_i \mid i\in[n]\}$}} 
\newcommand{\funcand}{\ensuremath{\textsc{And}_n}\xspace}       
\newcommand{\truthtable}{\ensuremath{\mathcal{K}}\xspace}       

\newcommand{\xorproblem}{$\textsc{Xor}_n$\xspace}               
\newcommand{\xorproblemparam}[2]{$\textsc{Xor}_n$-#1-#2\xspace} 
\newcommand{\xorproblemfull}{\xorproblemparam{$\{\textsc{Xor}\}$}{$\{x_i \mid i\in[n]\}$}} 
\newcommand{\funcxor}{\ensuremath{\textsc{Xor}_n}\xspace}       

\ifarxiv\else 
\fi

\newcommand{\toggleplot}[1]{{\textcolor{red}{Plots removed to increase compilation speed. Use command $\backslash$toggleplot in preamble to reinsert them.}}}

\iffinal
\newcommand{\andre}[1]{}
\newcommand{\romann}[1]{}
\newcommand{\cuong}[1]{}
\newcommand{\newedit}[1]{#1}
\newcommand*{\todo}[1]{}
\else
\newcommand{\andre}[1]{\textcolor{green!50!black}{[(Andre) #1]}}
\newcommand{\romann}[1]{\textcolor{purple}{[(Roman) #1]}}
\newcommand{\cuong}[1]{\textcolor{orange}{[(Cuong) #1]}}
\newcommand{\newedit}[1]{\textcolor{blue}{#1}}
\newcommand*{\todo}[1]{\textcolor{red}{\textrm{(TODO: #1)}}}
\fi

\ifreview
\usepackage[switch,mathlines]{lineno}

\usepackage{etoolbox} 
\newcommand*\linenomathpatch[1]{
  \cspreto{#1}{\linenomath}%
  \cspreto{#1*}{\linenomath}%
  \csappto{end#1}{\endlinenomath}%
  \csappto{end#1*}{\endlinenomath}%
}
\linenomathpatch{equation}
\linenomathpatch{gather}
\linenomathpatch{multline}
\linenomathpatch{align}
\linenomathpatch{alignat}
\linenomathpatch{flalign}
\fi

\begin{document}

\ifarxiv
\author[1]{Duc-Cuong~Dang}
\author[2]{Roman~Kalkreuth}
\author[1]{Andre~Opris}

\affil[1]{University of Passau, Passau, Germany}
\affil[2]{RWTH Aachen University, Aachen, Germany}
\date{}
\fi

\title{Runtime Analysis of Cartesian Genetic Programming in Evolving Boolean Functions}
\maketitle

\begin{abstract}
%
Cartesian Genetic Programming (CGP) 
is among the practical and popular forms of Genetic Programming as it uses a graph-based representation of programs. 
%
%
This paper presents a first runtime analysis of CGP in 
evolving Boolean functions \newedit{using complete training sets}. 
We prove \newedit{an} asymptotic bound \newedit{$O(nD^5)$} for the expected 
\newedit{number of fitness evaluations} 
of CGP to construct a conjunction of $n$ inputs using at most $D\geq n-1$ 
binary gates, a minimal function set\newedit{, and even with a strict survival selection.  
When the non-strict selection is used, the bound is improved to $O(nD^4)$.} 
Our analysis reveals interesting characteristics of CGP induced search, 
which have been only observed empirically\newedit{. In particular,} enabling 
the acceptance of equally good solutions\newedit{, including those} with connected gates 
non-contributing to fitness, 
can lead to a speedup, and consequently a better 
asymptotic time bound. 
In contrast \newedit{to conjunctions}, 
we also prove a negative result 
which shows that CGP requires exponential time to evolve an exclusive disjunction. 
\newedit{Experiments evolving conjunctions complement our theoretical findings. The use of incomplete training sets is found to further reduce the average number of fitness evaluations while maintaining a good level of generalisation.}
\end{abstract}

\ifreview
\linenumbers
\fi

\section{Introduction}\label{sec:intro}

Genetic Programming (GP) is a search paradigm originally proposed for the automated discovery of computer programs. 
GP leverages evolutionary search using selection and variation to evolve programs given a specification, constraints, and input-output examples. 
Early work by Cramer~\cite{Cramer1985} and Dickmanns et al.~\cite{DickmannsSW1987} applied Genetic Algorithms (GA) to evolve computer programs. 
However, Koza~\cite{Koza1989,Koza1993} later introduced a tree-based form of GP to evolve symbolic expressions, commonly implemented as LISP S-expressions at that time, which widely popularised the concept of GP. 
The traditional form of GP therefore uses tree data structures to model programs, and this is commonly referred to as Tree-based GP (TGP). 
The leaf nodes represent the inputs of the program and are assigned elements from a \emph{terminal set} $T$, which consists of variables or constants. The root and internal nodes compute functions from a \emph{function set} $F$, taking the arguments from the outputs of their respective child nodes. After evaluation, the value produced at the root node constitutes the final output of the modeled program. Over time, GP has been extended to incorporate alternative program representations, including linear sequences~\cite{Perkis1994,Openshaw94buildingnew}, graph-based structures~\cite{Poli1996,10.5555/2934046.2934074}, and rule-based models~\cite{RyanCO1998}. 

In this work, we are interested in Cartesian Genetic Programming (CGP)~\cite{MillerT2000}, which 
is, unlike TGP, a graph-based form of GP.
%
CGP is a well established methodology 
in logic synthesis and evolvable hardware~\cite{DBLP:journals/tec/VasicekS15},
with many successful applications such as in designing 
    high order combinational circuits~\cite{DBLP:conf/eurogp/Vasicek15}, 
    complex approximate circuits~\cite{DBLP:journals/gpem/VasicekS16}, 
    and cryptographic functions~\cite{DBLP:conf/cec/HusaS20}. 
These successes are believed to be 
    due to the high versatility of the graph-based representation. 
%
Learning Boolean functions was also among the motivating applications for the introduction of CGP~\cite{10.5555/2934046.2934074}. 
%
%
However, it remains poorly understood when and why even simple CGP algorithms perform well in
this problem domain, and 
%
a solid theoretical foundation for CGP in evolving Boolean functions is 
still missing. 
%
%

\textbf{Our contribution:} 
Inspired by the work of Mambrini and Oliveto~\cite{MambriniO2016} on TGP, we initiate a rigorous analysis of CGP for evolving Boolean functions by analysing the runtime of the \opocgp and the strict variant \opocgpstrict. We consider the popular single active-gene mutation (SAM)~\cite{GoldmanP2013} and the minimal function sets. 
For constructing conjunctions of $n$ variables with $D \geq n-1$ \textsc{And} gates and the complete training set, we obtain expected runtimes of $\bigO(n D^4)$ and $\bigO(n D^5)$, respectively, using fitness-level arguments and Markov chain analysis. The faster performance of \opocgp \newedit{over \opocgpstrict} stems from its non strict selection, which accepts equally fit solutions with redundant active gates, a beneficial effect previously observed only empirically~\cite{DBLP:conf/eurogp/TurnerM14}. 
%
%
Prior \newedit{counterpart} TGP results \cite{LissovoiO2018,MambriniO2016} make no distinction between strict and non strict selection, and in comparison, CGP is slower but guarantees to only use  $\bigO(D \log D)$ space. 
\newedit{The original CGP~\cite{10.5555/2934046.2934074} was not designed for speeding up TGP, but to offer a more general representation of programs 
with natural support for Boolean circuit applications. A CGP program in the Boolean domain is a logic circuit.
The aim of our paper is not about showing a superior performance, but rather to understand how and when polynomial performance guarantees can be achieved.}

In contrast \newedit{to conjunctions,} we prove that \opocgp requires exponential time with exponentially high probability, to construct an exclusive disjunction of $n$ inputs out of $(1+d)n\geq D\geq n-1$ \textsc{XOR} gates for some $d<1$ and using the complete training set. Here we apply the negative drift theorem~\cite{OlivetoW2011}. 
Experiments on conjunctions support the theory, showing clear gains by using non strict selection. 
\newedit{We also conduct experiments using incomplete static training sets, and observe that this setting further reduces the number of fitness evaluations required by CGP while maintaining a good level of generalisation.}

\textbf{Related work:} 
From the empirical side, 
several 
studies have examined search behaviour of CGP, primarly focusing on characteristics such as neutral search~\cite{10.1007/3-540-45355-5_16}, redundancy~\cite{DBLP:journals/tec/MillerS06}, bloat~\cite{DBLP:conf/eurogp/TurnerM14} and evolvability in the fitness landscapes 
induced by logic synthesis problems~\cite{10.1007/3-540-45355-5_16}. Moreover, based on the results of empirical studies, it has been demonstrated and  hypothesised that CGP-driven program search benefits from high levels of redundancy, enabled by the use of very large genotypes~\cite{DBLP:journals/tec/MillerS06,DBLP:conf/eurogp/TurnerM14}. This also includes handling neutral fitness landscapes by neutral genetic drift (NGD), which has been shown to help escape local optima~\cite{10.1007/3-540-45355-5_16,DBLP:journals/gpem/TurnerM15a}. 
A step forward in understanding the complexity characteristics of CGP induced search in logic synthesis was made by the proposal of the General Boolean Function Suite (GBFS)~\cite{10.1145/3594805.3607131}. This is a diverse benchmark 
of different types of Boolean functions
commonly used to evaluate GP methods. 

Mambrini~and~Oliveto~\cite{MambriniO2016} were the first to rigorously analyse GP 
in evolving Boolean functions by considering the same TGP systems as~\cite{NeumannOW2011,KoetzingSNO2014} using the minimal function sets. 
On the complete training set, they showed that the \opotgp and \opotgpstrict with the single-step (local) mutation constructs a conjunction of $n$ inputs in $\bigO(n\log{n})$ expected fitness evaluations. On a static incomplete training sampled uniformly at random of $s\in\poly(n)$ rows, the time bound to fit the training set is reduced to $\bigO(\log{n})$. Using the negative drift theorem, they also showed that the \opotgp requires exponential time to construct an exclusive disjunction of $n$ variables. Other setting were also considered in their paper, such as constructing exclusive disjunction with incomplete training set, and conjunction with negated input variables. 
These results were later extended for the algorithms with the multi-step (global) mutation \newedit{in~Lissovoi~and~Oliveto~\cite{LissovoiO2018,LissovoiO2019},} which added the generalisation ability guarantee for conjunctions on incomplete training sets, along with many other results. \newedit{Doerr~et~al.~\cite{DoerrLO2019,DoerrLO2023}} pushed this line of research further by considering \newedit{the settings of} non-minimal function sets, \newedit{unknown target function size,} and introduced a new tool, called the super-multiplicative theorem, for their analysis. 
 
For CGP, we are unaware of any theoretical research in learning Boolean functions. 
The only work on runtime analysis of CGP that we know of is from Kalkreuth~and~Droschinsky~\cite{KalkreuthD2019} on 
one function node.
\newedit{Finally, the book chapter~\cite{LissovoiO2020} summarises the current state-of-the-art theoretical research on GP.}

\section{Preliminaries}\label{sec:prelim}

%
We use the 
asymptotic notation with symbols $\Omega$, $\bigO$,
$\Theta$~\cite{Cormen2009}.
The sets of integer and natural numbers are $\Z$ and $\N$ respectively.
For $n \in \N$, define $[n] \coloneqq \{1,\dots,n\}$ and $[n]_0 \coloneqq [n] \cup \{0\}$.
The natural logarithm is denoted $\ln(\cdot)$ and that of base-2 is $\log(\cdot)$.
We use $H_n$ to denote the $n$-th harmonic number, \ie 
$\ln(n+1) < H_n\coloneqq\sum_{i=1}^{n}\frac{1}{i} < 1+\ln(n)$, 
thus $H_n = \bigO(\log{n})$.
%
\newedit{Let $\mathrm{Li}_s(z) := \sum_{i=1}^{\infty}\frac{z^i}{i^s}$ be 
the polylogarithm (or Jonqui\`{e}re's) function for $s,z\in\mathbb{C}$, 
\ie extended by analytic continuation. Some special cases are 
$\mathrm{Li}_{-2}(z)=\sum_{i=1}^{\infty}i^2 z^i = \frac{z(1+z)}{(1-z)^3}$
for $|z|<1$, and $\mathrm{Li}_{s}(1)=\sum_{i=1}^{\infty}\frac{1}{i^s}
=:\zeta(s)$ which is known as the Riemann's Zeta function. Note that 
$\zeta(2)=\pi^2/6$ and $\zeta(4)=\pi^4/90$.}
The logic \textsc{And}, \textsc{Or} and exclusive \textsc{Or} are denoted 
$\wedge$, $\vee$ and $\XOR$ respectively. 
%
%
The uniform distribution over a set $A$ is denoted $\Unif(A)$ and 
the geometric distribution with parameter $p\in[0,1]$ is denoted $\Geom(p)$. 
For two random variables $X,Y$, $X\succeq Y$ means that $X$ stochastically
dominates $Y$, that is, for any $a \in \mathbb{R}$ it holds $\prob{X\leq a} 
\leq \prob{Y\leq a}$. 

\subsection{Cartesian Genetic Programming}\label{sec:cgp}

CGP is capable of modelling programs with $\Nin$ inputs and $\Nout$ outputs, and uses 
$\Nfunc$ (internal) function nodes. The functions that can be used to compute the output of those nodes are 
defined in a lookup table of $\Kfunc$ entries indexed from $0$ to $\Kfunc-1$. 
The function specified in each entry has an arity of at most $\arity$. 
A (valid) program for such a CGP is then represented by a directed acyclic graph $G=(V,E)$,
referred to as the \emph{phenotype}, which defines how the outputs are computed by composing functions of the lookup table and using the respective inputs. 
The inputs and the lookup table model the same concepts as the terminal 
set $\termset$ and the function set $\funcset$ in Tree-based GP, thus any TGP program always has an equivalent CGP program. 

The original form of CGP~\cite{10.5555/2934046.2934074} was proposed to arrange the function nodes on a grid-like 
structure similar to an FPGA circuit. However, we consider the modern (simplified) approach 
in which the nodes are arranged in a single line, thus $V$ can be indexed from $0$ to $\Nall-1$ 
where $\Nall:=\Nin + \Nfunc + \Nout$ or simply $V=[\Nall-1]_0$, \ie starting with 
the input nodes $\Nin$, then the function nodes $\Nfunc$, and finally the output ones $\Nout$. 
Each input node has no incoming edges (or connections) while each output node has exactly 
one coming from a non-output node, \ie with index in $[\Nall-\Nout-1]_0$. 
Each function node is labelled with a function of the lookup table and has exactly
$\arity$ incoming edges from nodes of strictly smaller indices, thus from the inputs
or from the outputs of the preceding function nodes. These edges represent the sequence of 
arguments that the function at the node takes, thus there is an ordering among them. 
If the arity of the corresponding function in the lookup table is smaller $k$, then only some 
first edges matter.

\begin{figure}[t]\centering
\begin{tikzpicture}[scale=0.95,every node/.style={scale=0.95}]\small 
\newcommand\nodeindex[1]{\scriptsize\textcolor{blue}{#1}}
\tikzset{middlearrow/.style={
                decoration={markings,
                mark= at position 0.5 with {\arrow{#1}} ,
            }, postaction={decorate} } }
\tikzstyle{block}=[rectangle, draw, rounded corners, minimum height=0.6cm]
\tikzstyle{node}=[circle, draw, fill=white, minimum size=0.65cm]
\tikzstyle{edge}=[]
\tikzstyle{branch}=[fill,shape=circle,minimum size=1em,inner sep=0pt,scale=0.2]

\begin{scope}[shift={(0,5.3)}] 
\node[anchor=west] (lt) at (-1.0,0) {\textbf{lookup table}};
\node[right of=lt, node distance=3.5cm] {\footnotesize
\begin{tabular}{c|c|l}
index & symbol & func.\\ \hline
$0$ & $\wedge$ & \textsc{And} \\
$1$ & $\vee$ & \textsc{Or} \\
$2$ & $\downarrow$ & \textsc{Nor} \\
\end{tabular}
};
\end{scope}

\begin{scope}[shift={(-0.5,3.0)}] 
\node at (3,-1.3) {\textbf{phenotype}};
\node[node,rectangle,label=-180:{\nodeindex{$0$}}] (x0) at (0,1)   {$x_1$};
\node[node,rectangle,label=-180:{\nodeindex{$1$}}] (x1) at (0,0)  {$x_2$};
\node[node,rectangle,label=-180:{\nodeindex{$2$}}] (x2) at (0,-1)  {$x_3$};
\node[node,label=90:{\nodeindex{$3$}}]  (n3) at (1.2,0) {$\wedge$};
\node[node,label=90:{\nodeindex{$4$}}]  (n4) at (2.4,0) {$\downarrow$};
\node[node,label=90:{\nodeindex{$5$}}]  (n5) at (3.6,0) {$\vee$};
\node[node,label=-90:{\nodeindex{$6$}}] (n6) at (4.8,0) {$\downarrow$};
\node[node,label=90:{\nodeindex{$7$}}] (n7) at (6.0,0) {$\wedge$};
\node[node,diamond,label=90:{\nodeindex{$8$}}] (n8) at (7.2,0) {$o_1$};

\draw[edge] (x0) edge[bend left,middlearrow={>}]  (n3);
\draw[edge] (x1) edge[bend right,middlearrow={>}] (n3);
\draw[edge] (n3) edge[bend left,middlearrow={>}]  (n4);
\draw[edge] (x2) edge[bend right,middlearrow={>}] (n4);
\draw[edge] (n4) edge[bend left,middlearrow={>}]  (n5);
\draw[edge] (n3) edge[bend right,middlearrow={>}] (n5);
\draw[edge] (x0) edge[bend left,middlearrow={>}]  (n6);
\draw[edge] (n4) edge[bend right,middlearrow={>}] (n6);
\draw[edge] (n5) edge[bend left,middlearrow={>}]  (n7);
\draw[edge] (x1) edge[bend right,middlearrow={>}] (n7);
\draw[edge] (n7) edge[bend left,middlearrow={>}]  (n8);
\end{scope}

\begin{scope}[shift={(0,0.5)}] 
\node[anchor=west] (nnl) at (-1.0,0) {\textbf{genotype}};
\node[block,right of=nnl,label=90:{\nodeindex{$3$}},node distance=5em] (nn2) {$0$ \nodeindex{$0$} \nodeindex{$1$}};
\node[block,right of=nn2,label=90:{\nodeindex{$4$}}] (nn3) {$2$ \nodeindex{$3$} \nodeindex{$2$}};
\node[block,right of=nn3,label=90:{\nodeindex{$5$}}] (nn4) {$1$ \nodeindex{$4$} \nodeindex{$3$}};
\node[block,right of=nn4,label=90:{\nodeindex{$6$}}] (nn5) {$2$ \nodeindex{$0$} \nodeindex{$4$}};
\node[block,right of=nn5,label=90:{\nodeindex{$7$}}] (nn6) {$0$ \nodeindex{$5$} \nodeindex{$1$}};
\node[block,right of=nn6,label=90:{\nodeindex{$8$}},node distance=0.8cm] (nn7) {\nodeindex{$7$}};
\end{scope}

\begin{scope}[shift={(7.6,5.3)}] 
\node[anchor=west] (ecl) at (0.0,-3.6) {\textbf{equivalent circuit}};

\node (x1) at (0.5,0) {$x_1$};
\node (x2) at (0.0,0) {$x_2$};
\node (x3) at (1.0,0) {$x_3$};

\node[and gate US,draw,logic gate inputs=nn,scale=0.5] at ($(x3)+(0.5,-2.0)$) (n3) {};
\node at ($(n3)+(0,0.3)$) {\nodeindex{$3$}};

\node[nor gate US,draw,logic gate inputs=nn,scale=0.5] at ($(x3)+(1.0,-1.0)$) (n4) {};
\node at ($(n4)+(0,0.3)$) {\nodeindex{$4$}};

\node[or gate US, draw,logic gate inputs=nn,scale=0.5] at ($(x3)+(1.5,-2.0)$) (n5) {};
\node at ($(n5)+(0,0.3)$) {\nodeindex{$5$}};

\node[and gate US,draw,logic gate inputs=nn,scale=0.5] at ($(x3)+(2.0,-3.0)$) (n7) {};
\node at ($(n7)+(0,0.3)$) {\nodeindex{$7$}};

\draw (n7.output) -- ($(n7.output)+(0.5,0)$) node[above] {$o_1$};
\draw (x1) |- (n3.input 1);
\draw (x2 |- n3.input 2) node[branch] {} -- (n3.input 2);
\draw (x3) |- (n4.input 1);
\draw (n3.output) |- (n4.input 2);
\draw (n4.output) |- (n5.input 1);
\draw (n3.output) |- (n5.input 2);
\draw (n5.output) |- (n7.input 1);
\draw (x2) |- (n7.input 2);
\end{scope}
\end{tikzpicture}
\caption{Example of a CGP program that models a Boolean function.
         Node~6 is a so-called inactive node since no path touching this node leads to the output.}
\label{fig:example-cgp-program}
\end{figure}
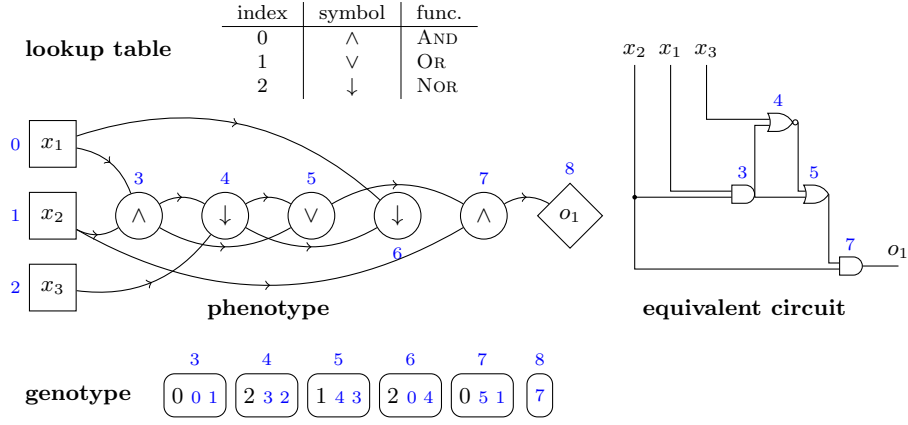

It is possible for a CGP program to ignore some of the function nodes, which are referred to
as \emph{inactive} (function) nodes. More generally, a non-terminal node of $V$ is called \emph{active} 
if there is a path from this specific node to one of the output nodes, otherwise it is called inactive. By this 
definition, output nodes are always active. 
The \emph{genotype} (or the encoding) of $G$ is a vector of 
$\Nfunc(\arity+1)+\Nout$  integers. 
The sequence is divided into $\Nfunc+\Nout$ blocks indexed from 
$\Nin$ to $\Nall-1$, and these indices match those nodes in the phenotype that the blocks encode. A block with index $i<\Nin+\Nfunc$ (a so called function block) contains
$\arity+1$ integers (or genes), the first, referred to as the \emph{function gene}, is a number 
in $[\Kfunc-1]_0$ (pointing to an index in the lookup table) while the rest, 
referred to as \emph{function connection genes}, are integers in $[i-1]_{0}$. 
Blocks with a larger index than $i$ (output blocks) only contain one integer in 
$[\Nall-\Nout-1]_0$, called the \emph{output connection gene}. 
Thus the set of all genotypes (all valid CGP programs), denoted by $\searchspace$, 
is an integer-valued space. Unlike TGP, CGP therefore has a fixed space complexity 
to store a program as stated in the lemma below. Figure~\ref{fig:example-cgp-program} gives 
an example of a CGP program with its phenotype representation and the corresponding genotype encoding. 

\begin{lemmarep}\label{lem:cgp-space-complexity}
Only $(\Nfunc(\arity+1)+\Nout)\log(1+\max\{\Kfunc, \Nin+\Nfunc\})$ storage space is required
to store a CGP program.
\end{lemmarep}
\begin{appendixproof}
The program is encoded by a vector of $\Nfunc(\arity+1)+\Nout$ integers. Each integer is 
bounded from above by $\max\{\Kfunc, \Nin+\Nfunc\}$, thus at most 
$\log(1+\max\{\Kfunc, \Nin+\Nfunc\})$ bits are required to encode it. The total number of 
bits is $(\Nfunc(\arity+1)+\Nout)\log(1+\max\{\Kfunc, \Nin+\Nfunc\})$.
\end{appendixproof}

\begin{algorithm2e}[b]\small
$x\sim \Unif(\searchspace)$\;
\For{$\mathrm{iter}:= 1 \to \infty$}{
    $y:=\mathrm{mut}(x)$\;
    \lIf{$f(y)\leq f(x)$}{\label{algo:one-plus-one-cgp:line:selection}
        $x:=y$
    } 
}
\caption{The ($1$+$1$)-CGP on the space $\searchspace$ of programs.}
\label{algo:one-plus-one-cgp}
\end{algorithm2e}

We consider the simplest CGP algorithm the so-called \opocgp (Algorithm~\ref{algo:one-plus-one-cgp} for minimisation). Starting from
a program initialised uniformly at random, in each iteration a new offspring program $y$ is 
generated by mutating the current one. The offspring will replace the parent only if it has better or 
equally good fitness. Here $f(x)$ is referred to as the fitness of a program $x$ 
which is problem-dependent and assumed given. 
By convention in runtime analysis, we let the algorithm run infinitely then estimate the expected time to reach some target state. 
%
Like TGP~\cite{MambriniO2016}, we also 
refer to the variant of the \opocgp where strict comparison is used instead in 
Line~\ref{algo:one-plus-one-cgp:line:selection} as the \opocgpstrict.

\begin{algorithm2e}[ht]\small
$b:=\lfloor i/(\arity+1)\rfloor$ \tcp{\scriptsize block index}
$j:=i \mod (\arity+1)$ \tcp{\scriptsize index inside a block}
\eIf{$i<\Nfunc(\arity+1)$}{
    \lIf{$j=0$}{
        $S:=[\Kfunc-1]_0$ \tcp*[h]{\scriptsize valid functions}
    }
    \lElse{
        $S:=[\Nin+b-1]_0$ \tcp*[h]{\scriptsize valid input connections}
    }
}{
    $S:=[\Nin+\Nfunc-1]_0$ \tcp{\scriptsize valid output connnections}
}
\If{$|S\setminus\{x[i]\}|\geq 1$}{
    $S:=S\setminus\{x[i]\}$ \tcp{\scriptsize avoid the old value if possible}
}
$v\sim \Unif(S)$\;
\KwRet{$v$}
\caption{$\textsc{NewValue}(x, i)$.}
\label{algo:new-value}
\end{algorithm2e}


\begin{algorithm2e}[ht]\small
Copy $x$ to $y$\;
$\mathrm{active} := \mathrm{false}$\;
\While{$\neg \mathrm{active}$}{
    $i\sim\Unif([\Nfunc(\arity+1)+\Nout-1]_0)$\;
    $y[i]:=\textsc{NewValue}(y, i)$\;
    $\mathrm{active} := \textsc{IsActive}(x,i) \wedge (y[i]\neq x[i])$\;
}
\KwRet{$y$}
\caption{$\textsc{MutationSAM}(x)$.}
\label{algo:mut-sam}
\end{algorithm2e}

\newedit{Mutations} 
in CGP can alter any gene (component) of the genotype (vector) to a new value 
uniformly at random while respecting the constraint on their range, this is done with the 
help of the procedure \textsc{NewValue} which is summarised in Algorithm~\ref{algo:new-value}.
When enforcing the change of the gene value is not possible, for example if 
only one function is specified in the lookup table then the function genes are fixed to zero 
and cannot be mutated, 
the old value is returned instead.

%
\newedit{The original paper~\cite{Miller1999} proposed \emph{point mutation} operators 
in which a fixed or random number of genes (mutation points) are chosen uniformly at random for alternation.}
\newedit{Nowadays, the so-called \emph{single active-gene mutation} 
(SAM)~\cite{GoldmanP2013} is more popular and is our focus. 
SAM is} parameter-free and \newedit{is} described in Algorithm~\ref{algo:mut-sam}.
\newedit{It} 
repeatedly modifies genes chosen uniformly at random until 
an active node is modified. 
\newedit{T}he detection of active nodes can be done during the program evaluation\newedit{, so} 
we assume that there is a function $\textsc{IsActive}$ that can tell whether a given gene belongs 
to an active node or not in an evaluated program. 
%

\subsection{Learning Boolean Functions with GP}\label{sec:logic-synth}


Learning Boolean functions or, more generally, logic circuits
with heuristic methods is part of the wider field of logic synthesis~\cite{DBLP:books/daglib/0086041}. 
\newedit{In this section, we describe how the learning is evaluated for any GP.}
 
The problem of learning a Boolean function $g\colon \bool^n \to \bool$ on $n$ inputs
where $\bool:=\{\valtrue,\valfalse\}$ is to compose a program/circuit a set of elementary 
logic gates to exactly match or to approximates $g$ using 
a set $S$ of input-output examples of $g$ referred to as the \emph{training set}. 
To accommodate the construction of $S$ by random sampling, we rather consider $S$ as 
a multi-set, each of its elements is called a \emph{row} which consists of a pair of 
example input assignment $\alpha\in \bool^n$ and the desired output $g(\alpha)\in\bool$. 
Formally, the set of \newedit{input variables} is $\termset := \{\inputi{i} \mid i \in[n]\}$ and 
coincides with the GP terminal set. Then each $\alpha\colon \termset \to \bool$ is 
a mapping of every input to a Boolean value. Like $g$, a given candidate program $y$ is 
also a function $x_1,\ldots,x_n \mapsto y(x_1,\ldots,x_n)$ on the inputs. 
To calculated the fitness of $y$, the number of mismatched outputs of $y$ on $S$ is counted, 
that is, 
\[ f(y):=|\{(\alpha,g(\alpha))\in S 
            \mid 
            y(\alpha(\inputi{1}),\ldots,\alpha(\inputi{n})) 
                \neq g(\alpha(\inputi{1}),\ldots,\alpha(\inputi{n}))\}|.
\]
Thus the smaller $f(y)$ is the closer $y$ is to fully fit the training set $S$.

In the 
so-called synthesis \emph{with the complete training set}, the 
complete specification of $g$ is available and is used to evaluate candidate
programs, that is, $S=\truthtable_{g}$ where 
$\truthtable_{g}:=\{(\alpha,g(\alpha(\inputi{1}),\ldots,\alpha(\inputi{n})))\}_{\alpha\in\textsc{bool}^n}$ 
is the complete truth table of $g$. 
Since the truth table grows exponentially in $n$, \ie $|\truthtable_{g}|=2^n$, 
in practice to reduce the computational time 
\newedit{of} each fitness evaluation  
\newedit{for large $n$}, 
only a fraction of $\truthtable_{g}$ is sampled and used as training 
set $S$. This is referred to as synthesis \emph{with an incomplete training set}. 
We 
assume that $S$ is sampled once and remains unchanged 
\newedit{during evolution}, 
\ie $S$ is 
called a \emph{static} training set. 

\newedit{We only consider the complete training set in our theoretical analysis, the
reason is discussed in the last section of the paper. However, in experiments, incomplete 
training sets are considered.} 
Similarly to~\cite{DoerrLO2019,MambriniO2016}, we define the \emph{generalisation error} 
$\varepsilon_g(y)$ of a program $y$ in modelling a target function $g$ \newedit{as} 
the probability that $y$ returns a mismatched output of a row selected uniformly at random 
from 
$\truthtable_{g}$, \ie
$\varepsilon_g(y):=\prob{y(\alpha(\inputi{1}),\ldots,\alpha(\inputi{n}))\neq g(\alpha(\inputi{1}),\ldots,\alpha(\inputi{n}))}$
where $\alpha\sim \Unif(\bool^n)$. The \emph{generalisation ability} of $y$ is the 
complement of its generalisation error, \ie $G_g(y):=1-\varepsilon_g(y)$. Thus $y$ fully 
meets the specification of $g$ if and only if $G_g(y)=1$. A value of $G_g(y)$ close $1$ 
means that $y$ generalises $g$ well while a value close to $1/2$ means that it is only 
as good as a random guess. A program $y$ with $G_g(y)$ close to $0$ can be turned to 
$\bar{y}$ with $G_g(\bar{y})$ close to $1$, thus $y$ and $\bar{y}$ are equally informative 
of $g$. 

\newedit{The exact computation of $G_g(y)$ is of course expensive, however it can
be approximated by considering a \emph{validation set} (also called \emph{test set}) $S'$ 
which is sufficiently large and is created independently from $S$ by sampling the rows 
of $\truthtable_{g}$ uniformly at random with replacement. The fitness of $y$ evaluated
on $S'$, for which $y$ was not evolved to fit, gives an estimation of $\varepsilon_g(y)$ on $S'$, 
and is denoted $\hat{\varepsilon}(y,S')$. The quantity $\hat{G}(y,S'):=1-\hat{\varepsilon}(y,S')/|S'|$
is the maximum likelihood estimator, and hence an approximation, of $G_g(y)$.}

\section{Analysis of Evolving Conjunctions}\label{sec:analysis-and}
The first target Boolean function that we \newedit{analyse} 
is the
\andproblem function of $n\geq 2$ inputs, \ie $\funcand(x_1,\ldots,x_n) =
\inputi{1} \wedge \inputi{2} \wedge \ldots \wedge \inputi{n}$, that must be
built from at least $n-1$ binary $\funcbinand$ gates. 
\newedit{I}n GP terminology, we
have the function set $F=\{\funcbinand\}$ and the terminal set $T:=\{x_i\mid
i\in[n]\}$. Similar to the naming convention for the \maxproblem
problem~\cite{KoetzingSNO2014}, but here without restricting the tree depth as
it is not part of the problem definition, we adapt the name \andproblemfull for
this logic synthesis problem. Regarding the complete training set of this function, only one row of $\truthtable_{\funcand}$
has $\funcand(\alpha(\inputi{1}),\ldots,\alpha(\inputi{n}))=\valtrue$ when
$\alpha(\inputi{1})=\ldots=\alpha(\inputi{n})=\valtrue$, all the other rows have
the desired output $\valfalse$.

To implement this synthesis in CGP, it suffices to set $\Nin = n$, $\Nout = 1$, use a lookup table with the single binary function $\textsc{And}$ (i.e., $\Kfunc = 1$ and $\arity = 2$), and choose $\Nfunc \in \mathbb{N}$ such that $\Nfunc \geq n - 1$.
%
%
It follows from \cite{MambriniO2016} that the fitness of programs for \andproblemfull is 
monotonically improved (here decreased for the minimisation problem) as a function of the number 
of \newedit{input variables used/included (connected to active nodes in case of CGP).}

\begin{lemma}[Lemma~1 in~\cite{MambriniO2016}]\label{lem:and-connecting-i-var}
On \andproblemfull, the truth table of a program currently using $i$ distinct \newedit{input variables} 
differs from the target truth table of $x_1 \wedge x_2 \wedge \ldots 
\wedge x_n$ at exactly $f_i = 2^{n-i} - 1$ rows.
\end{lemma}

We show that \opocgp and \opocgpstrict with $\Nfunc := D\geq n-1$, using SAM mutation and the complete training set, synthesi\newedit{s}e the target function of \andproblemfull in expected time polynomial in $D$. In particular, for $D \in [n-1,\infty)\cap O(n)$, the expected time is $\bigO(n^6)$ and the space required is $\bigO(n\log n)$.

\begin{theoremrep}\label{thm:and-cgp-complete-eval}
The \opocgp and \opocgpstrict fit the complete training set 
of \andproblemfull using 
    $\Nfunc:=D\geq n-1$, 
    and SAM mutation 
in $\bigO\left(nD^5\right)$ fitness evaluations in expectation 
and uses $\bigO(D\log D)$ space to store its programs. 
If $D\geq 10$, the probability that the number of fitness evaluations exceeds 
$(1+\delta)20\pi^2 (n-1) D^5$ for any $\delta\geq 1$ is at most $e^{-\delta/3}$.
\end{theoremrep}
\begin{proofsketch}
From Lemma~\ref{lem:and-connecting-i-var} and the monotonicity of $2^{n-i}-1$ in $i$, 
\newedit{the number of input variables included in the current program cannot be decreased in future iterations.} 
Hence, we apply the fitness-level method~\cite{Wegener2002}, using as levels the number of included input variables, to bound the expected time until all variables are included. When estimating the probability $s_i$ to leave the current level to a higher one, 
we only consider strict improvements, thus the results hold for both \opocgp and \opocgpstrict. 
There are two cases, either there is a free (input) connection from an active node that can redirected
to a new input variable, otherwise there are none. The former case gives a higher probability
of $\frac{1}{3D+1}\cdot \frac{n-i}{D+n-1}$. In the latter case, the SAM mutation 
needs to re-arrange the connections of a non-active node so as to extend to a new input 
variable and to make sure that when the node is activated, the fitness is improved. 
The probability in this case is at least $\frac{2\cdot \alpha (n-i) \cdot (n-i)}{(3D+1)^3(2D)^3}=
\frac{\alpha (n-i)^2}{(3D+1)^3(4D^3)}$ where $\alpha:=D/(n-1)$. 
The expected running time is then bounded from above by 
\begin{align*}
\sum_{i=1}^{n-1}\frac{(3D+1)^3(4D^3)}{\alpha (n-i)^2} 
    &=(3D+1)^3 4 D^3 \alpha^{-1}\sum_{i=1}^{n-1}\frac{1}{i^2}
     \leq (3D+1)^3 4 D^3 \alpha^{-1} \zeta(2)
\end{align*}
and this is 
$(3D+1)^3 D^3 ((n-1)/D) (2\pi^2/3) =\bigO\left(n D^5\right)$.
The tail bound then follows by applying Theorem 1 of Witt~\cite{Witt2014}.
\end{proofsketch}
\begin{appendixproof}
We apply the fitness level method to bound the expected running time.  
The search space $\searchspace$ of valid programs is partitioned into $n$ fitness levels and
the ($1$+$1$)~CGP is said to be at level $i$ if its current program has $i$ inputs connected 
to the output via its active function nodes. It follows from Lemma~\ref{lem:and-connecting-i-var}, 
and because $2^{n-i}-1$ is a monotone function, that the algorithm cannot fall down to a level 
below its current level. We then estimate for each level $i\in[n-1]$ a lower bound $s_i$ on 
the probability of leaving the current level to a higher one. 

Our arguments will only consider strict improvement steps thus the implied result holds for 
both \opocgp and \opocgpstrict. Another way to have $D \geq n-1$ is to assume that $D=\alpha(n-1)$ 
for some $\alpha\geq 1$.

For a function node to be 
active, either an input edge of another function node (after it) or the output connection 
must be connected to it. So, for a program with $i$ active function nodes at least $i-1$ 
input edges of these nodes have been already used to maintain the node activeness, 
here the $-1$ takes into account the output connection. This leaves at most $2i - (i - 1) 
= i+1$ of the remaining input edges of these nodes to connect to the inputs, in other words
the program can be at most at level $i+1$. Conversely, a program at level $i$ must have at 
least $i-1$ active function nodes. We consider the following cases.

\textbf{Case 1}: 
If there is at least one input edge of an active function node that can be redirected 
to an unused input, we then estimate the probability of this redirection in one iteration of 
the while-loop in SAM. The component associated with that input edge is selected for modification 
with probability $\frac{1}{3D+1}$, then the modification points the edge to one of 
the $n-i$ unused inputs with probability $\frac{n-i}{D+n-1}\geq \frac{n-i}{2D}$. Thus, 
the probability of leaving the current level in this case is at least 
$
\frac{n-i}{(3D+1)(2D)}
    = \frac{(n-i)(D-i+1)}{(3D+1)(2D)(D-i+1)}
    \geq  \frac{\alpha(n-i)^2}{(3D+1)(2D^2)}
$. The last inequality holds because 
$D-i+1\leq D$ and 
$D-i+1
    = \alpha(n-1)-i+1 
    = \alpha n - (\alpha i - \alpha i + \alpha+i-1)
    = \alpha n - (\alpha i - (\alpha-1)(i-1))
    \geq \alpha n - \alpha i$ for $i\geq 1$ and $\alpha\geq 1$. 

\textbf{Case 2}: Otherwise, if no input edges of the active nodes can be redirected to 
an unused input to increase the current level, then there are exactly $i-1$ active function 
nodes and at least $D-(i-1)\geq \alpha(n-i)$  
inactive function nodes. We then estimate the probability to activate one of these
nodes and leave the current level within the three first iterations of the while-loop in SAM.  
In the first two iterations, we make some changes to the input edges of an inactive node 
by assuming the worst case that these changes are required.
In the last iteration, we activate the node and this ends the while-loop in SAM. 

The probability to make a change to an input edge of an inactive node in the first iteration 
of the while-loop is at least $\frac{2\cdot \alpha(n-i)}{3D+1}$. Then this input edge is redirected 
to an unused input $x_u$ with a probability of at least $\frac{n-i}{D+n-1}\geq \frac{n-i}{2D}$. 
Let $j$ be the inactive node associated with the input edge, our argument
on what may happen in the two last iterations depends on the location of $j$.

\textbf{Subcase 2.1}: Node $j$ is located before the first (leftmost) active function node or 
there are no active function nodes at all. Let $x_v$ be the input that one input edge 
$e$ of the first active function node is currently connected to, or that the output connection 
$e$ is currently connected to if there is no active function nodes at all. In the second 
iteration, the algorithm redirects the other input edge of $j$ (the one not selected in the 
first iteration) to $x_v$, with a probability of at least 
$\frac{1}{3D+1}\cdot \frac{1}{D+n-1} \geq \frac{1}{(3D+1)(2D)}$. 
To activate $j$ in the third iteration, it suffices to redirect edge $e$ to $j$, 
and the probability of this event is also at least $\frac{1}{(3D+1)(2D)}$. 
By considering all three iterations, the overall probability to leave the current level $i$ 
by SAM for this subcase is at least 
$
\frac{2\cdot \alpha (n-i) \cdot (n-i)}{(3D+1)^3(2D)^3}=
\frac{\alpha (n-i)^2}{(3D+1)^3(4D^3)}$. 

\textbf{Subcase 2.2}: Node $j$ is located after the first active function node. In the 
second iteration, we redirect the other input edge of $j$ to the active function node $j^{-}$ 
that is located immediately before $j$, with a probability of at least $\frac{1}{(3D+1)(2D)}$. 
In the last iteration, we redirect the connection $e$ (which can be the output one) that 
currently maintains the activeness of $j^{-}$ to node $j$ in order to activate $j$, 
with a probability of at least $\frac{1}{(3D+1)(2D)}$. 
The overall probability to leave the current level $i$ for this subcase is also at least 
$\frac{\alpha (n-i)^2}{(3D+1)^3(4D^3)}$. 

Combining all the cases gives that the probability to leave the current level $i$ is at least 
$s_i:=\frac{\alpha (n-i)^2}{(3D+1)^3(4D^3)}$ and summing their reciprocals over all the levels implies 
an expected running time of at most 
\begin{align*}
\sum_{i=1}^{n-1}\frac{(3D+1)^3(4D^3)}{\alpha (n-i)^2} 
    &=(3D+1)^3 4 D^3 \alpha^{-1}\sum_{i=1}^{n-1}\frac{1}{i^2}
     \leq (3D+1)^3 4 D^3 \alpha^{-1} \zeta(2)
\end{align*}
and this is 
$(3D+1)^3 D^3 ((n-1)/D) (2\pi^2/3) =\bigO\left(n D^5\right)$. The space complexity follows 
by applying Lemma~\ref{lem:cgp-space-complexity} with $\Nfunc=D, k=2$ and $\Nin=n=\bigO(D)$. 

For $D\geq 10$, we have that $(3D+1)^3 \leq (3D+D/10)^3 < 30D^3$, the expected running time can be upper bounded by $20\pi^2 (n-1) D^5$. Applying Lemma~\ref{lem:tail-bound-geometric} with 
$p_i:=\frac{\alpha (n-i)^2}{(3D+1)^3(4D^3)}$
thus $p_{\min}=\frac{\alpha}{(3D+1)^3(4D^3)}> \frac{D/(n-1)}{120D^6} = \frac{1}{120 (n-1) D^5}$ 
and
\begin{align*}
\sum_{i=1}^{n-1}\frac{1}{p^2_i} 
    &= (3D+1)^6(16D^6)\alpha^{-2}\sum_{i=1}^{n-1} \frac{1}{(n-i)^4} 
     \leq (30 D^3)^2(16 D^6) \alpha^{-2} \sum_{i=1}^{n-1} \frac{1}{i^4}\\
    &\leq 14400 \cdot D^{12} \cdot ((n-1)/D)^2 \cdot \zeta(4) 
     = 14400 \cdot D^{10} (n-1)^2 (\pi^4/90)\\ 
    &= 160\pi^4 (n-1)^2 D^{10} =: s,
\end{align*}
and with $\gamma=20\pi^2 \delta (n-1)D^5$, 
implies that the probability of exceeding the running time $(1+\delta)20\pi^2 (n-1) D^5$ 
is at most 
\begin{align*}
\exp\left(-\min\left\{\frac{400\pi^4 \delta^2 (n-1)^2 D^{10}}{4\cdot 160\pi^4 (n-1)^2 D^{10}}, 
                      \frac{20\pi^2 \delta (n-1) D^5}{4\cdot 120 (n-1) D^5}\right\}\right)
\leq e^{-\frac{\pi^2 \min\{\delta,\delta^2\}}{24}},
\end{align*}
and since $\min\{\delta,\delta^2\}=\delta$ for $\delta\geq 1$, this is less than $e^{-\frac{\delta}{3}}$.
\end{appendixproof}

\newedit{
The bound $\bigO(n^6)$ of the minimal setting $D=n-1$} is tight and 
\newedit{non-improvable  
for the \opocgpstrict when the initial program almost already fits the training set.}

\begin{theoremrep}\label{thm:and-cgp-complete-eval-strict-lower-bound}
\newedit{There exist initial programs for which the \opocgpstrict with 
    $\Nfunc:=n-1$, 
    and SAM mutation
requires $\Omega(n^6)$ fitness evaluations in expectation to fit the complete training 
set of \andproblemfull.}
\end{theoremrep}
\begin{appendixproof}
\newedit{The initial programs correspond to those obtained at the penultimate level, 
each having a near-optimal fitness value of $1$. A concrete example is the following. 
The first $n-2$ function nodes are the only active ones and are connected to
some $n-1$ distinct input variables. This means the program output is currently pointed 
to the $(n-2)$-th function node $j$. Meanwhile neither the remaining free variable $v$ 
nor node $j$ is connected to the last (inactive) function node $j^+$.}

\newedit{Modifying any input connection of the first $n-2$ nodes is not possible
since this triggers termination of the while-loop in SAM and the produced
solution is rejected by the strict selection as its fitness is at least $1$.
Modifying the connection of $j^+$ without creating an optimal program is also
impossible. Therefore, SAM mutation is forced to produce an optimal solution in
one call by redirecting the two input connections of $j^+$ to $j$ and $v$, then
by activating $j^+$ (this triggers the termination of the while-loop). Such an
event requires SAM to run for some $i\geq 3$ iterations in which the first $i-1$
iterations only modify the components of node $j^+$, with probability
$\left(\frac{3}{3(n-1)+1}\right)^{i-1}=\left(\frac{3}{3n-2}\right)^{i-1}$, 
and the last iteration redirects the program output to $j^+$, with probability
$\frac{1}{3(n-1)+1}\cdot \frac{1}{n+(n-1)-1}=\frac{1}{(3n-2)(2n-2)}$.
Conditioned on this, it still requires that among the first $i-1$
iterations that at least one iteration redirects one input connection of $j^+$ 
to $v$, and another redirects the other input connection to $j$. The conditional
probability for this by a union bound is at most ${i-1 \choose
2}\left(\frac{2}{3}\cdot\frac{1}{2n-2}\right)^2$ because two out of the three 
components associated with $j^+$ are for its input connections, and the probability
of a successful redirection to a specific node is $1/(2n-2)$.}

\newedit{Summing the probability over all possible $i$ gives the upper bound for 
the event, here we use $(i-1)/2\leq i-2$ that holds for $i\geq 3$:  
\begin{align*}
& \sum_{i=3}^{\infty} \left(\frac{3}{3n-2}\right)^{i-1}\cdot \frac{1}{(3n-2)(2n-2)}\cdot\frac{(i-1)(i-2)}{2}\left(\frac{2}{3}\cdot\frac{1}{2n-2}\right)^2\\
&\quad\leq \bigO(n^{-4}) \sum_{i=3}^{\infty} (i-2)^2\left(\frac{3}{3n-2}\right)^{i-2} \left(\frac{3}{3n-2}\right)\\
&\quad= \bigO(n^{-5}) \sum_{i=1}^{\infty} i^2\left(\frac{3}{3n-2}\right)^{i} 
      = \bigO(n^{-5}) \cdot \mathrm{Li}_{-2}\left(\frac{3}{3n-2}\right)\\
&\quad = \bigO(n^{-5}) \cdot \frac{(3/(3n-2))(1 + 3/(3n-2))}{(1 - 3/(3n-2))^3}
      = \bigO(n^{-6}), 
\end{align*} 
Hence, the expected time for the \opocgpstrict to fit the complete training set 
is $\Omega(n^6)$.} 
\end{appendixproof}

\newedit{In contrast, the upper} bound for the expected running time of \opocgp can be 
further 
\newedit{improved} by considering neutrality in the fitness landscape as follows. We 
need the following lemma on a three-state Markov chain. 
\newedit{The same chain was used in~\cite{CorusO2018,Oliveto2018} to analyse 
population diversity in Genetic Algorithms.}

\begin{lemmarep}\label{lem:markov-chain-three-states}
Consider the following Markov chain with three states with the absorbing state $3$ and 
parameters $p,q,r,s\in[0,1)$ where $q,r>0$.
\begin{center}
\begin{tikzpicture}[scale=1.0,every node/.style={scale=1.0}]\small
\tikzstyle{state}=[circle, draw, fill=white, minimum size=0.7cm]
\tikzstyle{absorb}=[circle, double, draw, fill=white, minimum size=0.7cm]
\tikzstyle{transit}=[-stealth]

\node[state]  (s1) at (0,0) {$1$}; 
\node[state]  (s2) at (2,0) {$2$}; 
\node[absorb] (s3) at (4,0) {$3$}; 

\draw[transit] (s1) edge[bend left] node[above] {$p$} (s3);
\draw[transit] (s1) edge node[above] {$q$} (s2);
\draw[transit] (s2) edge node[above] {$r$} (s3);
\draw[transit] (s2) edge[bend left] node[below] {$s$} (s1);

\draw[transit] (s1) edge[loop left] node[left] {$1-p-q$} (s1);
\draw[transit] (s2) edge[loop below] node[below] {$1-r-s$} (s2);
\end{tikzpicture}
\end{center}
The expected time $E_1$ to reach the absorbing state $3$ starting from state $1$ is
\[
E_1 
    = \frac{q+r+s}{rq+rp+sp} 
    \leq \frac{1}{r} + \frac{1}{q} + \frac{s}{rq}.
\]
Moreover, if we have $(p+q \geq q') \wedge (r\geq r') \wedge (s\leq s')$, then the above 
upper bound also holds with the bounding estimates, that is, 
$E_1 \leq \frac{1}{r'} + \frac{1}{q'} + \frac{s'}{r'q'}$. 
\end{lemmarep}
\begin{appendixproof}
Let $E_i$ be the expected time to reach the absorbing state $3$ starting from state $i$, then it
is clear that $E_3=0$. By the law of total probability, the Markov chain satisfies the system 
of equations: 
\begin{align}
    E_1 &= 1 + q E_2 + (1-p-q) E_1\label{eq:markov-chain-1}\\
    E_2 &= 1 + s E_1 + (1-r-s) E_2\label{eq:markov-chain-2}
\end{align}
From \eqref{eq:markov-chain-1}, we get 
$E_2 = \frac{(p+q)E_1 - 1}{q} = \left(\frac{p}{q}+1\right)E_1 - \frac{1}{q}$ and
substituting in \eqref{eq:markov-chain-2} gives
\begin{align*}
&s E_1 = \left(r+s\right)\left(\left(\frac{p}{q}+1\right)E_1-\frac{1}{q}\right) - 1\\
\Leftrightarrow &
\frac{r+s}{q}+1 = \left(\left(r+s\right)\left(\frac{p}{q}+1\right)-s\right) E_1.
\end{align*}
Therefore $E_1$ is
\begin{align*}
E_1 
    &= \frac{\frac{r+s}{q}+1}{\left(r+s\right)\left(\frac{p}{q}+1\right)-s}
     = \frac{\frac{(r+s)+q}{q}}{\frac{(r+s)(p+q) - sq}{q}}
     = \frac{q+r+s}{rq+rp+sp}\\
    &= \frac{(q+r+s)/(rq)}{(rq+rp+sp)/(rq)}
     = \frac{\frac{1}{r}+\frac{1}{q} + \frac{s}{rq}}{1 + \frac{p}{q} + \frac{sp}{rq}}
     \leq \frac{1}{r}+\frac{1}{q} + \frac{s}{rq}
\end{align*}
as $1 + \frac{p}{q} + \frac{sp}{rq}\geq 1$ and the first upper bound statement follows. 
To show the second bound, we start again from $E_1$ but divide both the numerator and 
denominator by $r(p+q)$ (instead of $rq$ in the previous calculation). This gives
\begin{align*}
E_1 &= \frac{(q+r+s)/(r(p+q))}{(rq+rp+sp)/(r(p+q))}
     = \frac{\frac{q}{r(p+q)}+\frac{1}{p+q} + \frac{s}{r(p+q)}}{1 + \frac{sp}{r(p+q)}}
     \leq \frac{1}{r'}+\frac{1}{q'} + \frac{s'}{r'q'}
\end{align*}
since $1 + \frac{sp}{r(p+q)}\geq 1$, $\frac{q}{p+q}\leq 1$, and 
$p+q\geq q'$, $r\geq r'$ and $s\leq s'$.
\end{appendixproof}

\begin{theoremrep}\label{thm:and-cgp-complete-eval-neutral-move}
The \opocgp fits the complete training set of 
\andproblemfull using 
    $\Nfunc:=D\geq n-1$, 
    SAM mutation
in $\bigO(nD^4)$ fitness evaluations in expectation. 
\end{theoremrep}
\begin{proofsketch}
Similarly to the proof of Theorem~\ref{thm:and-cgp-complete-eval}, we divide 
the search space into $n$ levels according to the number of input variables included.
But instead of estimating the probability to leave a level, we compute the expected
waiting time $\tau_i$ for that same event to happen so that we can apply Lemma~\ref{lem:markov-chain-three-states}. 

Therefore, we further divide each level $i$ into two sublevels $i^{+}$ and $i^{-}$ 
which match exactly 
\newedit{the two cases in the proof of Theorem~\ref{thm:and-cgp-complete-eval}, 
\ie there is at least a free connection versus there is none.}  
The lemma is then applied for each level $i$ with 
states $1,2,3$ being levels $i^{-},i^{+}, i+1$ respectively, thus $E_1$ gives an 
upper bound on $\tau_i$. Additionally, the probability of transiting from $i^{+}$ to $i+1$
is exactly the probability of leaving the level in \newedit{easier case,} 
thus at least 
$\frac{n-i}{(3D+1)(n+D-1)}\geq \frac{n-i}{(3D+1)(2D)}=:r'(i)$. The probability of leaving 
$i^{-}$ is 
\newedit{also higher compared that of the harder case of the previous proof} 
as it suffices to include a new function node without 
the requirement on increasing the fitness,
thus $p(i)+q(i)\geq \frac{2\alpha(n-i)}{(3D+1)^2(2D)^2} \eqqcolon q'(i)$. 
Finally, we argue that there are at most two function nodes that can be deactivated to 
return to $i^{-}$ from $i^{+}$, thus $s(i)\leq \frac{2}{3D+1}=:s'(i)$. Summing all $\tau_i$
gives the expected running time: 
\begin{align*}
\sum_{i=1}^{n-1} \tau_i 
  \leq& \sum_{i=1}^{n-1}\frac{1}{r'(i)} + \frac{1}{q'(i)} + \frac{s'(i)}{r'(i)q'(i)}\\ 
  \leq & \left((3D+1)(2D)+(3D+1)^2(2D)^2 (2\alpha)^{-1}\right) H_{n-1}\\ 
      &  + 2(3D+1)^2(2D)^3 (2\alpha)^{-1}\zeta(2). 
\end{align*}
This is $\bigO(n D^4)$, and here we use $\alpha=D/(n-1)$ as in the previous proof.
\end{proofsketch}
\begin{appendixproof}
Similarly to the proof of Theorem~\ref{thm:and-cgp-complete-eval}, we divide 
the search space into $n$ levels according to the number of distinct inputs used, and the \opocgp 
is said to be at level $i\in[n]$ if its current program is connected to $i$ distinct inputs. 
Again by Lemma~\ref{lem:and-connecting-i-var} and by the elitist selection, it holds that once 
the algorithm has reached level $i$ it cannot fall down to a level below $i$. We then estimate
the expected time $\tau_i$ to leave the current level $i$ to a higher level. 

To estimate of $\tau_i$ more precisely, by considering non-strict improvement steps as this is 
allowed by the \opocgp, we partition each level $i$ into two sublevels $i^{+}$ and $i^{-}$
which match exactly Cases~1 and~2 in the previous proof. That is, the algorithm is said to be 
at level $i^{+}$ (Case~1) if it has at least one input edge of an active node that can be 
redirected to an unused input, and it is at level $i^{-}$ (Case~2) otherwise. We then apply 
Lemma~\ref{lem:markov-chain-three-states} with states $1,2, 3$ being levels $i^{-},i^{+}, i+1$ 
respectively, and $\tau_i$ is bounded from above by $E_1$ of the lemma.

It remains to bound the parameters $p(i)+q(i),r(i),s(i)$ so that we can deduce an upper bound 
on $E_1(i)$. It follows from the same argument as in Case~1 of the previous proof that 
$r(i)\geq \frac{n-i}{(3D+1)(2D)}=:r'(i)$. 

To bound $p(i)+q(i)$ from below, we follow Case~2, its subcases and use the same notation 
per subcase and the same assumption but make slightly different changes within only two iterations
of SAM mutation because we only need to reach level $i^{+}$ (and not necessarily $i+1$) from $i^{-}$. 
In Subcase~2.1, the probability of selecting an input edge of an active function node $j$
for change in the first iteration is still at least $\frac{2\alpha(n-i)}{3D+1}$, but then it 
suffices to redirect that edge to $x_v$  thus with probability $\frac{1}{D+n-1}\geq \frac{1}{2D}$, 
then in the second iteration to redirect edge $e$ (either of the output or of the first active 
function node) to $j$, with a probability of at least $\frac{1}{(3D+1)(2D)}$.
In Subcase~2.2, we redirect the selected input edge of the function node $j$ to the first active
node $j^{-}$ before $j$ in the first iteration with probability $\frac{1}{D+n-1}\geq \frac{1}{2D}$, 
then redirect edge $e$ to $j$ in the second iteration with a probability of at least $\frac{1}{(3D+1)(2D)}$.
Therefore, we have that the probability of transiting from state $i^{-}$ to at least $i^{+}$ is 
$p(i)+q(i)\geq \frac{2\alpha(n-i)}{(3D+1)^2(2D)^2} \eqqcolon q'(i)$.

To bound $s(i)$ from above, we note that for the algorithm to be in level $i^{+}$ its current 
program must either have at least one input with multiple input connections from active function 
nodes to it, or have at least one active function node whose activeness is maintained by multiple 
input connections from other active function nodes. In the former case, for an upper bound argument on $s(i)$, we only need to focus one such input 
and two specific connections from the active nodes to the input because to return to level $i^{-}$ 
at least one of the connections must no longer come from an active node. In other words, there are 
at most two active nodes that can be deactivated/changed to return to level $i^{-}$, and 
this happens with a probability of at most $2\cdot\frac{1}{3D+1}$ by one SAM mutation. The argument is similar for the later case as again we only need to focus on one active function 
node and two specific connections that maintain the node activeness. Therefore, we have 
$s(i)\leq \frac{2}{3D+1}=:s'(i)$. 

Applying Lemme~\ref{lem:markov-chain-three-states} and summing all $\tau_i$ give the 
expected running time: 
\begin{align*}
\sum_{i=1}^{n-1} \tau_i 
  \leq& \sum_{i=1}^{n-1}\frac{1}{r'(i)} + \frac{1}{q'(i)} + \frac{s'(i)}{r'(i)q'(i)} \\
   =  & \left((3D+1)(2D)+(3D+1)^2(2D)^2 (2\alpha)^{-1}\right) \sum_{i=1}^{n-1}\frac{1}{n-i}\\
      &  + 2(3D+1)^2(2D)^3 (2\alpha)^{-1}\sum_{i=1}^{n-1}\frac{1}{(n-i)^2}\\
   =  & \; \bigO\left(n D^3\right) H_{n-1}  + \bigO(nD^4)\zeta(2)
   =  \bigO(nD^4).
\end{align*}
\end{appendixproof}

The analyses from Theorems~\ref{thm:and-cgp-complete-eval} and~\ref{thm:and-cgp-complete-eval-neutral-move} show that using more than the minimum $n-1$ function nodes can benefit CGP by increasing the bounds $s_i$ in 
\newedit{the harder case, and also} \newedit{$q'(i)$ for} $p(i)+q(i)$ in the Markov chain. 

\section{Analysis of Evolving Exclusive Disjunctions}\label{sec:analysis-xor}

The second target 
function that we consider is the $\funcxor$ function 
of $n \geq 2$ inputs where $\funcxor(x_1, \ldots , x_n)\coloneqq x_1 \XOR x_2
\XOR \ldots \XOR x_n$ that must be built from at least $n-1$ binary $\funcbinxor$ 
gates. 
This function is of interest for us because 
\newedit{it is not evolvable~\cite{Valiant2009} even though it is} PAC-learnable~\cite{Valiant1984} 
thus a candidate problem to show a negative result for CGP, especially for an advanced operator like SAM. We adopt the name \xorproblemfull for this 
problem. 
%
A candidate CGP program is a function whose \emph{raw form} may contain an
input variable repeated multiple times, \eg $x_1 \XOR x_1 \XOR x_2$ where $x_1$ is
repeated twice. 
However, a variable that is repeated an even number of times is as if it is ignored,  
and it is repeated an odd number of times is as if it is repeated once. 
We refer to
the function form 
after applying these simplifications as the \emph{simplified form}.

For the \xorproblemfull and using the complete training set
$\truthtable_{\funcxor}$, the fitness of a candidate program $y$ can only take 
two possible values, that is, $f(y)=0$ if the simplified form of $y$ is exactly 
$\funcxor$, \ie every input appears an odd number of times in the raw form, 
otherwise $f(y)=2^{n-1}$, \ie $y$ is incorrect on a half of the rows
(and correct on the other half)~\cite{MambriniO2016}. 
This property implies that if the use of an incomplete training set leads to 
a program that ignores some inputs in the simplified form, then the generalisation 
ability of such a program is always $2^{n-1}/2^n=1/2$. This is as bad as a random 
guess, 
\newedit{and justifies our focus} 
on the setting of the complete training set
$S=\truthtable_{\funcxor}$.

%
We then expect CGP to perform a random walk in the space of programs 
until it realises an optimal program, \ie having the simplified form $\funcxor$. 
To show an exponential time on reaching such a program, we first show that it is 
unlikely that such a program can emerge from a fixed subset $L$ of function nodes.
%
For a lower bound argument, it suffices to focus on two necessary properties of 
an optimal program built out of $L$: (i) every input is connected to $L$ by an 
odd number of connections; (ii) the output of at least $n-2$ nodes of $L$ is 
connected to other nodes in $L$. The second property is required because an optimal
program has at least $n-1$ active function nodes, thus at least $n-2$ of those have 
their activeness maintained by $L$. Formally, an input variable $x_i$ is called an \emph{odd variable} with respect 
to $L$ if it is connected to $L$ by an odd number of connections, otherwise it is 
an \emph{even variable} (with respect to $L$, we often omit this part when $L$ is 
clear from the context). Thus we track the number of odd variables, 
denoted $V(L)$, and 
$U(L):=|\{\ell \in L \mid \text{the output of $\ell$ is connected to another node in $L$}\}|$ 
of the current program in CGP to reach their respective values $n$ and $n-2$. 


\begin{lemmarep}\label{lem:xor-init}
A randomly generated CGP program on \xorproblemfull using $\Nfunc:=D\geq n-1$ has 
$V(L)\leq 3n/4$ with probability $1-e^{-\Omega(n)}$, for any fixed set $L$ of 
function nodes with $m:=|L|\in\poly(n)$.
\end{lemmarep}
\begin{appendixproof}
We are only interested in the sampling steps where the $2m$ 
connections of $L$ are sampled to connect to an input variable because those are 
the only steps that can change $V(L)$. Let $V_j$ be $V(L)$ of the sampled program 
after the $j$-th step and we apply the negative drift theorem (Theorem~\ref{thm:negative-drift}) with potential 
$X_j:=n-V_j$. Note that $X_0 = n$ as no connections of $L$ have been sampled yet 
and zero is an even number, and that $|X_{j+1}-X_j|=1$ as the sampling toggles 
the parity of exactly one input variable. 

Each considered sampling step chooses a input variable $x_1, \ldots, x_n$ uniformly 
at random. Thus for $X_j \in [n/4,3n/8]$, there are at least $V_j=n-X_j\geq n-3n/8
=5n/8$ odd variables, which increase $X_j$ if one is chosen. Otherwise if an even 
variable is chosen then $X_j$ is decreased. Thus, we have a drift
\[
\expect{X_{j+1}-X_j \mid X_j \in [n/4,3n/8]} \geq 5/8-3/8 = 1/4.
\]
Applying Theorem~\ref{thm:negative-drift} with parameters $a = n/4$, $b = 3n/8$,
$r(\ell) = 1$, and $\delta = 1/4$, we obtain for $T=\min\{j \in \mathbb{N} \mid
X_j \leq n/4\}$ that $P(T \leq e^{\varepsilon n}) \leq e^{-\Omega(n)}$ for a
suitable constant $\varepsilon>0$.
Since $T>2m$ implies the final $V(L)$ of the initialisation is below $3n/4$, thus 
event $\{T>2m\}\subseteq \{\text{final }V(L)<3n/4\}$ and we get 
$
\prob{\text{final }V(L)<3n/4}
    \geq \prob{T>2m}=1-\prob{T\leq 2m}
    \geq 1-\prob{T\leq e^{\varepsilon n}}
    \geq 1-e^{-\Omega(n)}$ as $2m\leq e^{\varepsilon n}$ for sufficiently large $n$.
%
%
\end{appendixproof}

If an optimal program has not yet been found, then the \opocgp always 
accepts offspring programs and all modifications by every iteration of 
the while-loop across SAM mutations are taken into account. We refer to 
each of these iterations as an \emph{elementary step}. The following lemma, 
again using the negative drift theorem, shows that it is unlikely for a
program with the two necessary properties, let alone for an optimal program, 
to emerge during the application of an exponential number of 
elementary steps to a randomly generated program.

\begin{lemmarep}\label{lem:xor-elem-steps}
Starting from a randomly generated CGP program on \xorproblemfull using 
$\Nfunc:=D\geq n-1$, for a fixed set $L$ of its function nodes with $m:=|L|\leq (1+d)n$
for any $d\leq 10^{-2}$, the probability that a program with $U(L)\geq n-2$ 
and $V(L)=n$ emerges during the application of $e^{\varepsilon n}$ elementary steps 
of SAM for some constant $\varepsilon>0$ is at most $e^{-\Omega(n)}$.
\end{lemmarep}
\begin{proofsketch}
We apply the negative drift theorem, \ie Theorem~2 in~\cite{OlivetoW2012}, 
using potential $Y_j = 2n - 2 - U_j - V_j$ where $U_j$ is $\min(U(L),n-2)$ 
and $V_j$ is $V(L)$ after applying $j$ elementary steps to the initial solution.
It follows from Lemma~\ref{lem:xor-init}, that $V_0\leq 3n/4$ with probability 
$1-e^{-\Omega(n)}$, thus \newedit{$Y_0\geq 2n-2-(n-2)-3n/4 = n/4$} with that same probability. We estimate the drift $\expect{Y_{j+1}-Y_{j}\mid Y_j}$ for $Y_j\in [n/256,n/128]$. 
Three events are of interest: 
(i) redirecting a connection from an odd input variable 
    and to another odd input variable, \ie with probability $\frac{V_j}{2m}\cdot\frac{V_j-1}{n+m-1}
    > \frac{(127n/128)^2}{(2m)^2}
    > \frac{6 n^2}{25 D^2}$; 
(ii) choosing a connection linked to an even input variable for modification, \ie with probability
    at most $(n/128+2dn)/(2m)\leq 1/128+2d$;
(iii) redirect a connection not linked to an input variable to an even input variable
or to a function node that is not counted in $U(L)$ yet, \ie with probability 
    at most $(n/128 + dn)/n + n/(128n) \leq 1/64 + d$. 
The first event contributes to the drift of $2$, the two last contribute at most $3$.
The drift is $\expect{Y_{j+1}-Y_j \mid Y_j \in [\frac{n}{256}-2,\frac{n}{128}-2]} > \frac{1}{1+d}\left(\frac{12}{25(1+d)} - \frac{3(1+d)}{42} - 9(1+d)d\right)$
which at least $3/10$ after plugging $d=10^{-2}$ in. The results then follow from 
the negative drift theorem.  
\end{proofsketch}
\begin{appendixproof}
For $m<n-1$, the result is trivial as $V(L)$ cannot reach $n$, so we assume 
the opposite. Let $U_j$ and $V_j$ be $\min(U(L),n-2)$ and $V(L)$ respectively after 
applying $j$ elementary steps. We can also neglect all steps where $U_j$ and $V_j$ 
are unchanged as including those can only increase the number of required steps.

We apply the negative drift theorem with potential $Y_j:=2n-2-U_j-V_j$. 
In any single iteration, the configuration of at least $n-2$ input variables remains 
unchanged, so $|V_{j+1} - V_j| \le 2$. Since only one link is modified per iteration, 
it is clear that $|U_{j+1} - U_j| \le 1$. Therefore, $|Y_{j+1} - Y_j| \le 3$ by 
triangle inequality. Further, $V_j$ is bounded from above by $n$, and $U_j$ is 
by $n-2$. 

It follows from Lemma~\ref{lem:xor-init}, that $V_0\leq 3n/4$ with probability 
$1-e^{-\Omega(n)}$. Assuming this occurs then $Y_0\geq 2n-2-(n-2)-3n/4\geq n/4$, 
and we estimate the drift $E[Y_{j+1}-Y_{j}\mid Y_{j}]$ in the interval $Y_j \in [n/256-2, n/128-2]$. 

In this interval it holds that
\begin{align}
    U_j &= 2n-2-Y_j-V_j\geq 2n-2-(\frac{n}{128}-2)-n = \frac{127n}{128}\label{eq:lb-uj}\\
    V_j &= 2n-2-Y_j-U_j\geq 2n-2-(\frac{n}{128}-2)-(n-2) \geq \frac{127n}{128} + 2\label{eq:lb-vj}
\end{align}

To increase $Y_j$, it suffices to select a connection pointing to an odd variable
and reconnect it to another odd variable (\wrt $L$), \ie with probability 
at least
\[
\frac{V_j}{2m}\cdot\frac{V_j-1}{n+m-1}
    \overset{\eqref{eq:lb-vj}}{>} \frac{(127n/128)^2}{(2m)^2}
    > \frac{6 n^2}{25 D^2} \eqqcolon p_{+}.
\]
for $n$ sufficiently large. This increases $Y_t$ by two. 
There are also other possibilities to increase $Y_t$ 
but we 
assume they only contributes zero to the drift. 

To decrease $Y_j$, it is necessary that one of the following two events happens. 
Note that some subsequent change following those events may lead to eventually no 
change in $Y_j$, but we pessimistically assume that $Y_j$ always is decreased by $3$.
\begin{itemize}
\item[($A$)] Select a connection pointing an even variable (then reconnect it). 
\item[($B$)] Select a connection not pointing to an input variable and either reconnect 
it to an even variable, or reconnect it to a function node of $L$ which is not currently 
counted in $U(L)$.
\end{itemize}

There are at most $2D-U_j-V_j \leq 2n+2dn-U_j-V_j = 2dn + 2 + Y_j \leq
n/128+2dn$ connections pointing to an even variable, thus 
$\prob{A}\leq (n/128+2dn)/(2m)\leq 1/128+2d$. Here we use
$n/(2m)\leq n/(2(n-1))\leq 1$ for $n\geq 2$.

It follows from \eqref{eq:lb-uj} that at most $D-1-U_j\leq n+dn-1-127n/128 <
n/128 + dn$ function nodes have their output not used within $L$, and from
\eqref{eq:lb-vj} that there are at most $n-V_j\leq n-(127n/128-2) < n/128$ even
variables. Note that there are at least $n$ choices for reconnection,  so
$\prob{B}\leq (n/128 + dn)/n + n/(128n) \leq 1/64 + d$. Here we assume that the
probability of selecting the right connection type is at most $1$.

So one of the two events happens with probability at most 
\[
\prob{A}+\prob{B}\leq \frac{1}{128}+2d+\frac{1}{64}+d < \frac{1}{42}+3d\eqqcolon p_{-}.
\]

Since $Y_j$ can only decrease by at most three in each iteration, we obtain for
the drift for $n$ sufficiently large and any $d \leq 10^{-2}$ that 
\begin{align*}
&\expect{Y_{j+1}-Y_j \mid Y_j \in [\frac{n}{256}-2,\frac{n}{128}-2]} > 2\cdot p_{+}-3\cdot p_{-}
= \frac{12n^2}{25 D^2} - \frac{3}{42} - 9d \\
&= \frac{n}{D}\left(\frac{12n}{25D} - \frac{3D}{42n} - \frac{9Dd}{n}\right) 
 \geq \frac{1}{1+d}\left(\frac{12}{25(1+d)} - \frac{3(1+d)}{42} - 9(1+d)d\right) \\
&> \frac{1}{1+10^{-2}}\left(\frac{12}{25(1+10^{-2})} - \frac{3(1+10^{-2})}{42} - 9(1+10^{-2})10^{-2}\right)  
 > \frac{3}{10} \eqqcolon \delta.
\end{align*}

Applying Theorem~\ref{thm:negative-drift} with parameters $a=n/256-2$, $b=n/128-2$,
$r(\ell)=3$ and $\delta$ we obtain for $T:=\min\{j \in \mathbb{N} \mid Y_j \leq
a\}$ that $P(T \leq e^{-\varepsilon n}) \leq e^{-\Omega(n)}$ for a suitable
constant $d>0$. Hence, even including the event at initialisation,  
with probability at least $1 - e^{-\Omega(n)}$ 
for $n$ sufficiently large, we still have $Y_j> n/256$ and thus $U_j + V_j <2n-2$, or 
equivalently $(V_j<n) \vee (U_j<n-2)$
throughout the application of $e^{\varepsilon n}$ elementary steps.
\end{appendixproof}



To extend the lemma to \opocgp, we take a union bound over all subsets $L$, whose number can be estimated via the approximation of the standard binomial coefficient bounds using binary entropy. Finally, the SAM while-loop terminates once the output connection is selected for modification, which occurs with probability $1/(3D+1)=\Omega(1/n)$ for $D=\bigO(n)$. Hence, an exponential number of elementary SAM steps translates to the same order of SAM mutations (fitness evaluations) by a Chernoff bound.

\begin{theoremrep}
\newedit{Let $d<10^{-2}$ be a sufficiently small constant. The probability that the \opocgp fails to fit the complete training set of \xorproblemfull within $e^{\Omega(n)}$ fitness evaluations using SAM mutation and $\Nfunc:=D$ for $n-1 \leq D \leq n+dn$, is $1-e^{-\Omega(n)}$.}
\end{theoremrep}
\begin{appendixproof}
It follows from Lemma~\ref{lem:xor-elem-steps} that the probability that an optimal program emerges from a fixed subset $L$ of $m$ function nodes with $m\in[n-1,n+dn]$ is at most $e^{-\Omega(n)}\leq e^{-\kappa n}$ after applying $e^{\varepsilon n}$ elementary steps of SAM to a randomly generated program for some constants $\kappa>0$ and $\varepsilon>0$. 
By a union bound over all possible such subsets $L$, we get a required running time for the \opocgp in terms of number of elementary steps.  
Since we may assume that $D \leq n+dn \leq 2n-2$, we may estimate the number of such subsets by 
\[
\sum_{i=n-1}^D \binom{D}{i} \leq (D-n)\binom{D}{n-1} \leq (D-n)2^{D \cdot H((n-1)/D)}
\]
for $n$ sufficiently large where \(H(\alpha) = -\alpha \log \alpha - {(1-\alpha)} \log (1-\alpha)\) for $\alpha \in (0,1)$ denotes the binary entropy function. 
The last inequality uses 
$\binom{n}{k} \;\le\; 2^{\,n H(k/n)}$, 
which holds for all $k \in [n]$ (see for example~\cite{EntropyBinom}). Note that $H(\alpha)$ is strictly monotone decreasing in $\alpha \in [1/2,1)$, and that $(n-1)/D \geq 1/(1+d+o(1))$ and hence, $D \cdot H((n-1)/D) \leq n \cdot (1+d)\cdot H(1/(1+d+o(1))) < n\kappa'< n\kappa$ for a suitable constant $\kappa'>0$, 
thus the required running time is still exponential. 

To count this time in terms of fitness evaluations, 
we note that for a SAM mutation, its while-loop ends if the output node is chosen for modification, \ie with probability at least $1/(3D+1)=\Omega(1/n)$. Now consider a sum $X$ of $e^{dn}$ Bernoulli random variables, each becoming one if the output is selected for modification and zero otherwise. 
Then because a SAM mutation can also be stopped by modifying an active function node, $X$ is a lower bound for the number of fitness evaluations. 
By a classical Chernoff bound, we have that $\Pr(X \geq e^{cn}) = 1-e^{-\Omega(n)}$ for a sufficiently small constant $c>0$, 
and this concludes our proof.
\end{appendixproof}

\section{Experiments}\label{sec:experiments}

We evaluate CGP and TGP on \andproblem through two sets of experiments. First, using the complete training set, we measure runtime to highlight the performance gap between the two approaches when evolving the exact solution. Second, with incomplete training and testing sets, we assess both runtime and generalisation ability to examine their generalisation performance.\\

\textbf{Experimental setup}:  We used TinyverseGP~\cite{KalkreuthOJADVH2025}, a modular benchmarking Python library for GP that provides modules for various flavours of GP including CGP and TGP, to perform the experiments. 
Using the existing implementations \texttt{TinyTGP} and \texttt{TinyCGP} of 
the library, we derived 
the models \texttt{SimpleTGP} and \texttt{SimpleCGP}
which correspond to ($1$ + $1$)~TGP/CGP as described in Section~\ref{sec:cgp} 
and~\cite{KoetzingSNO2014,KoetzingSNO2012}. 
The CGP model uses \textsc{MutationSAM} as described in Algorithm~\ref{algo:mut-sam},
while the TGP model uses \newedit{the mutations from~\cite{KoetzingSNO2012} which are} based on the HVL-prime \newedit{operator~\cite{NeumannOW2011}.} 
The implementations of the models and the \andproblem as well as the raw data obtained in the experiments are provided in our repository\footnote{https://github.com/GPBench/TinyverseGP/tree/ppsn-2026}. 

 \begin{table}[t]
\centering 
\scalebox{\ifarxiv0.95\fi}{
\begin{tabular}{@{\extracolsep{4pt}}llll}
\toprule   
 \textbf{AND$_n$ param.} & \textbf{Setting} & 
 \textbf{GP param.} & \textbf{Setting} \\
\cmidrule{1-2}\cmidrule{3-4}
Dimension $n$ & $\{3 ,\ldots,15\}$ (compl.), $\{3 ,\ldots,50\}$ (incompl.) & Maximum iterations & $1 \times 10^6$\\
Input $i$ & $i \in  \{0,1\}$ & Search strategy & ($1$+$1$) \\ 
Function set & $ F = \{\funcbinand\}$ & Selection types & non-strict, strict\\
Terminal set & $T = \{x_1 ... x_n\}$ & No. replications per $n$ & 30\\ 
\bottomrule
\end{tabular}
}
\caption{Configuration of GP and \andproblem for the experiments.}
\label{tab:gp_configuration}
\end{table}

\begin{figure}[h]
\includegraphics[width=0.496\textwidth]{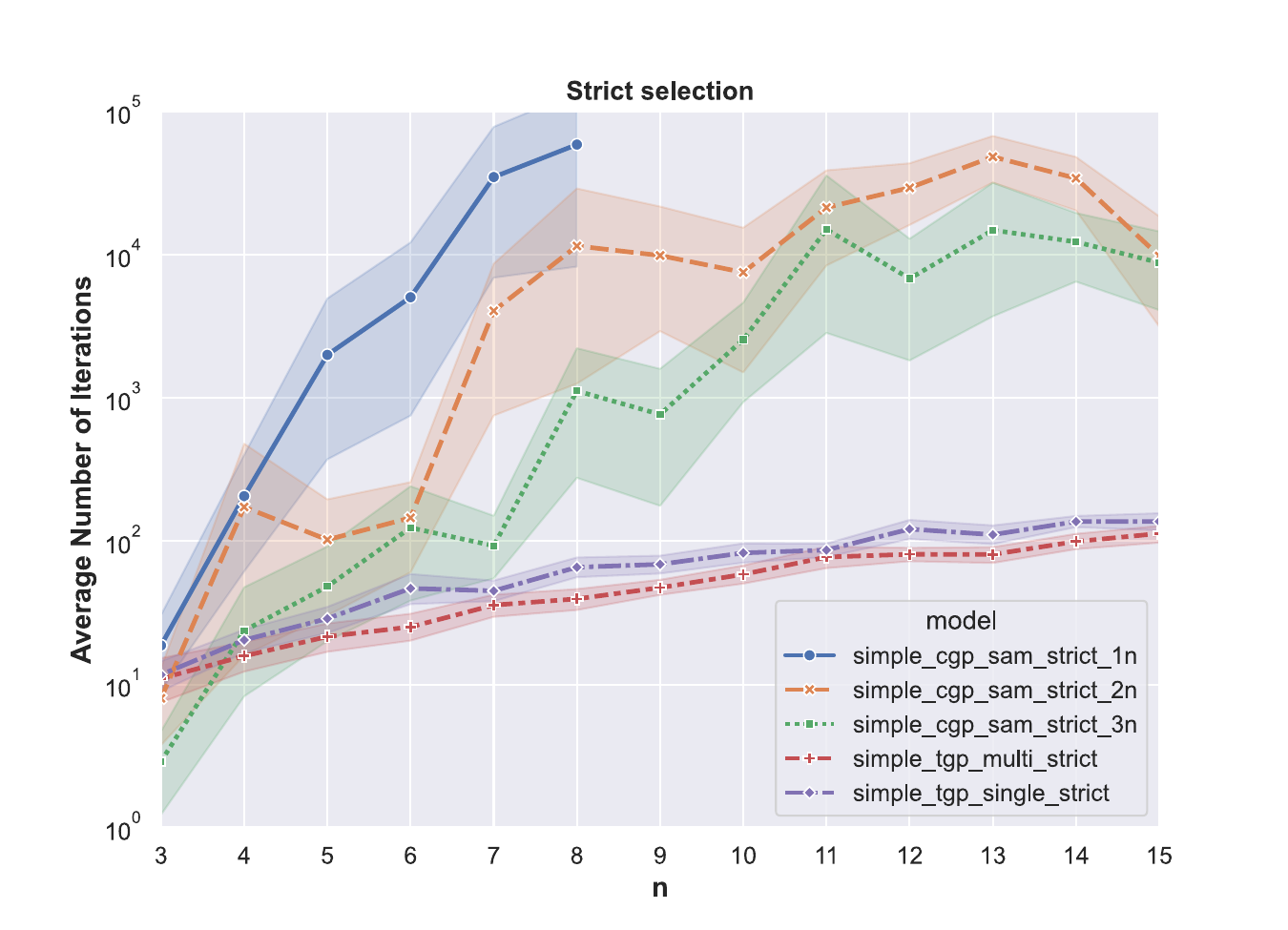} 
\hfill
\includegraphics[width=0.496\textwidth]{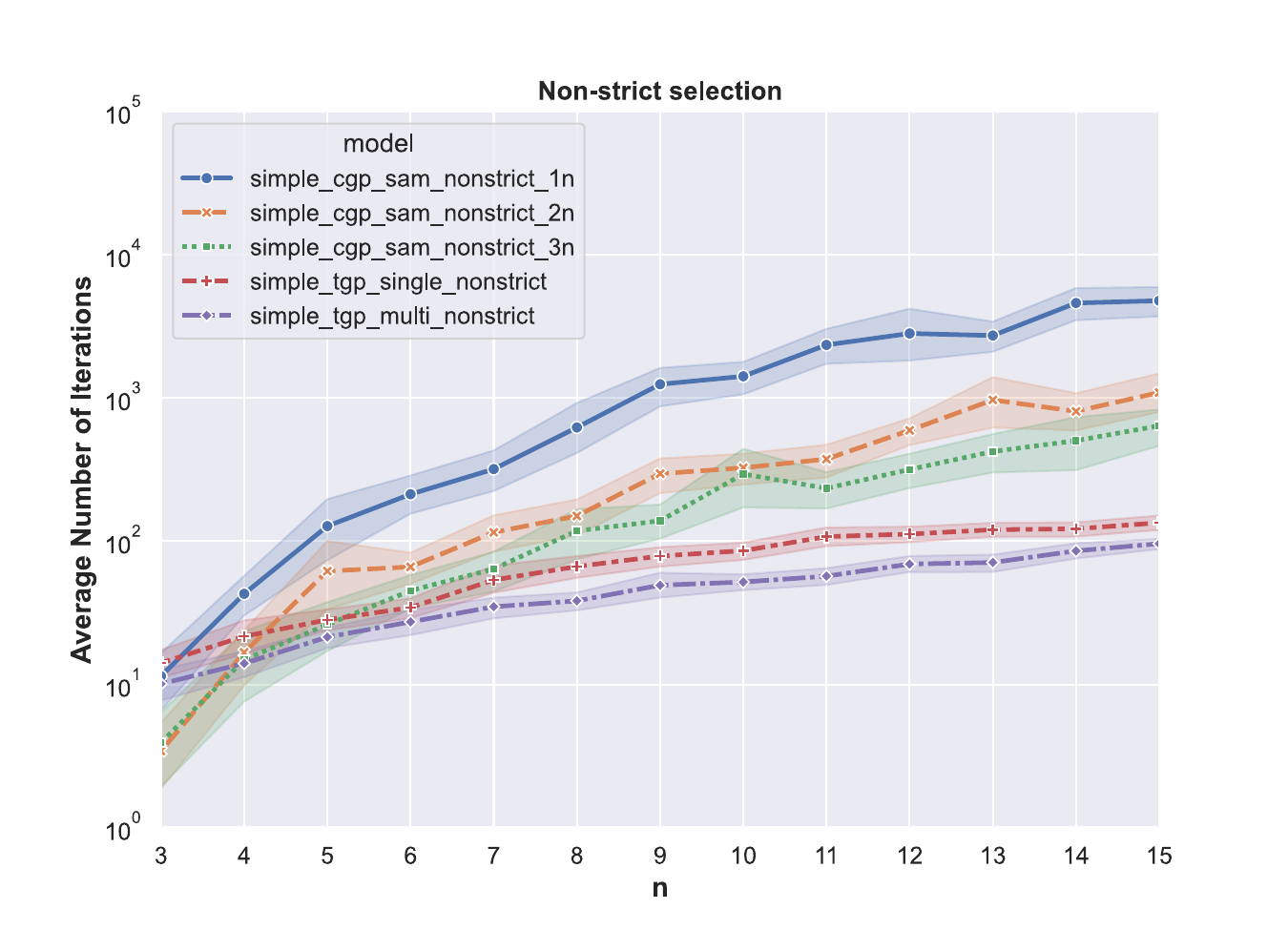} 
\caption{Results of the comparison on \andproblem using complete datasets.}
\label{fig:results-runtime-complete}
\end{figure}

\begin{figure}[h]
\includegraphics[width=0.496\textwidth]{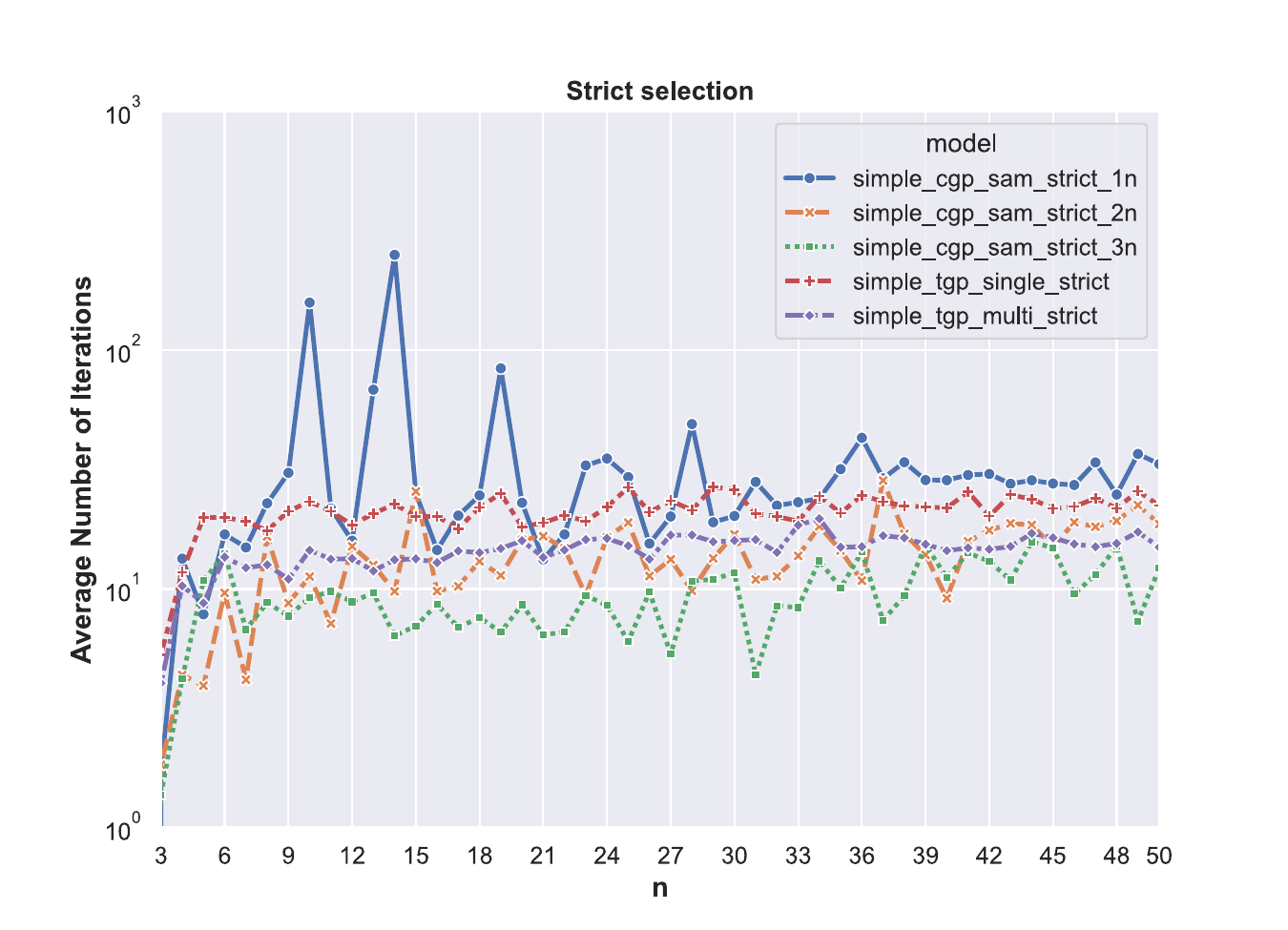} 
\hfill
\includegraphics[width=0.496\textwidth]{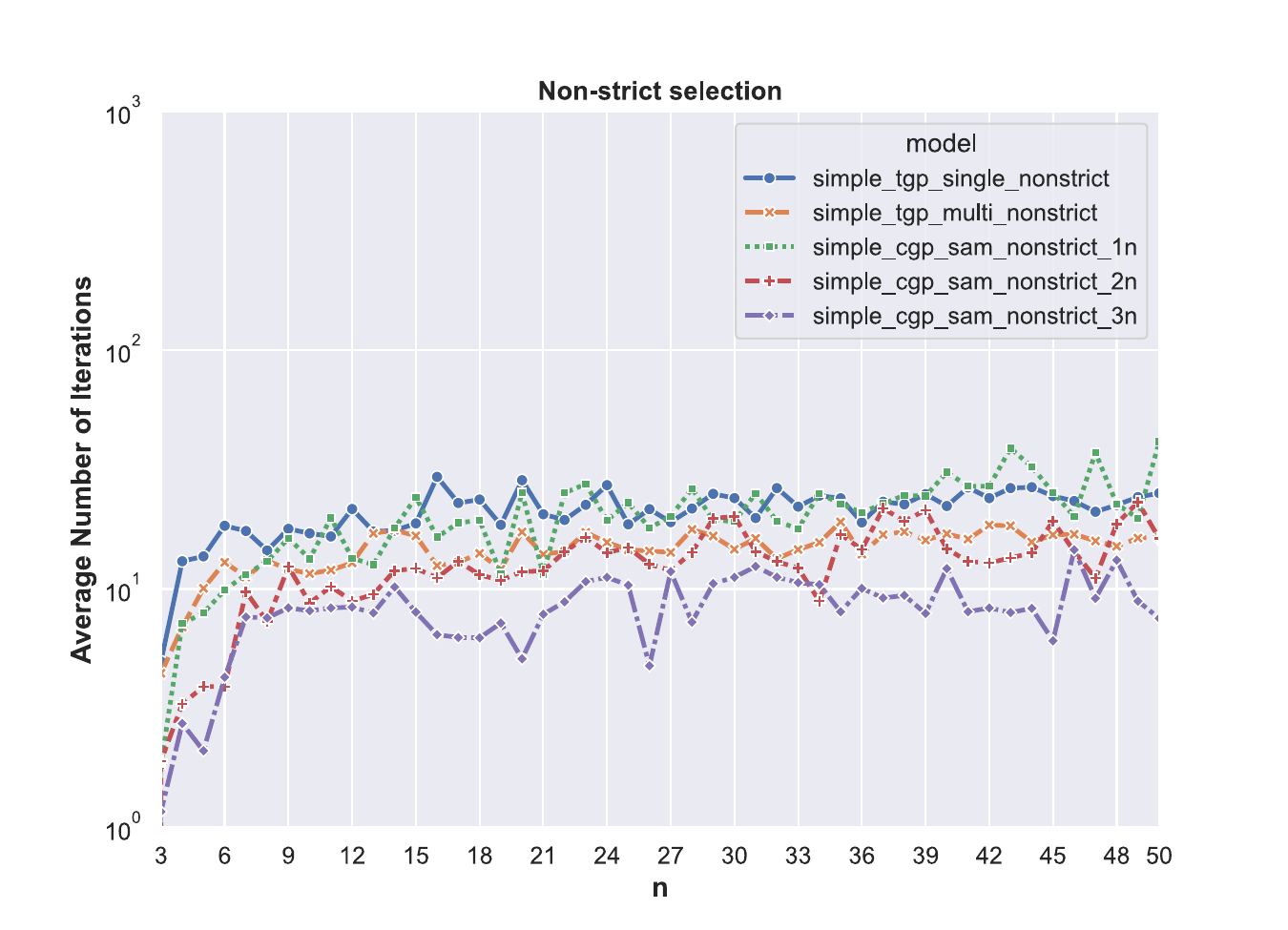} 

\includegraphics[width=0.496\textwidth]{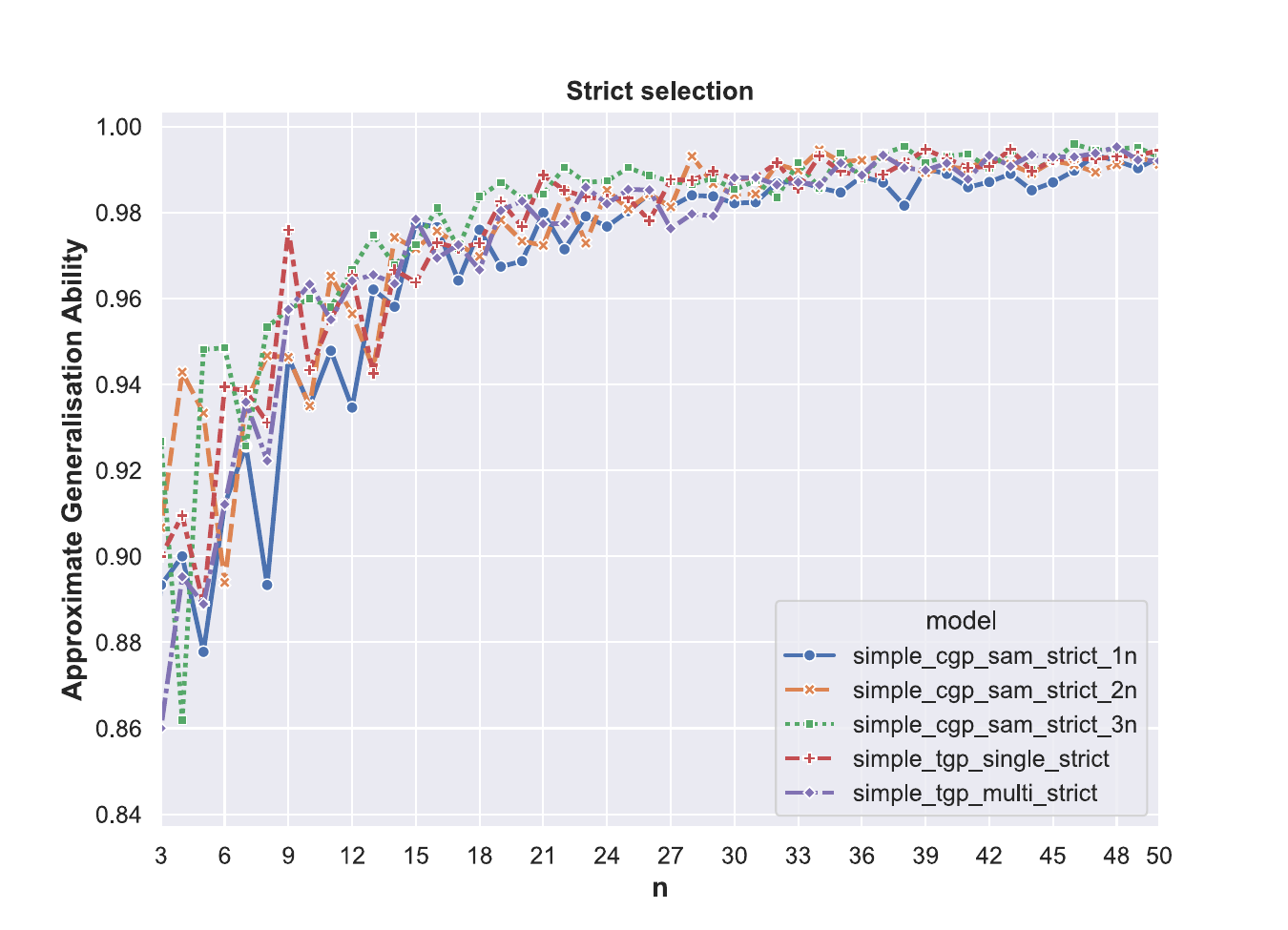} 
\hfill
\includegraphics[width=0.498\textwidth]{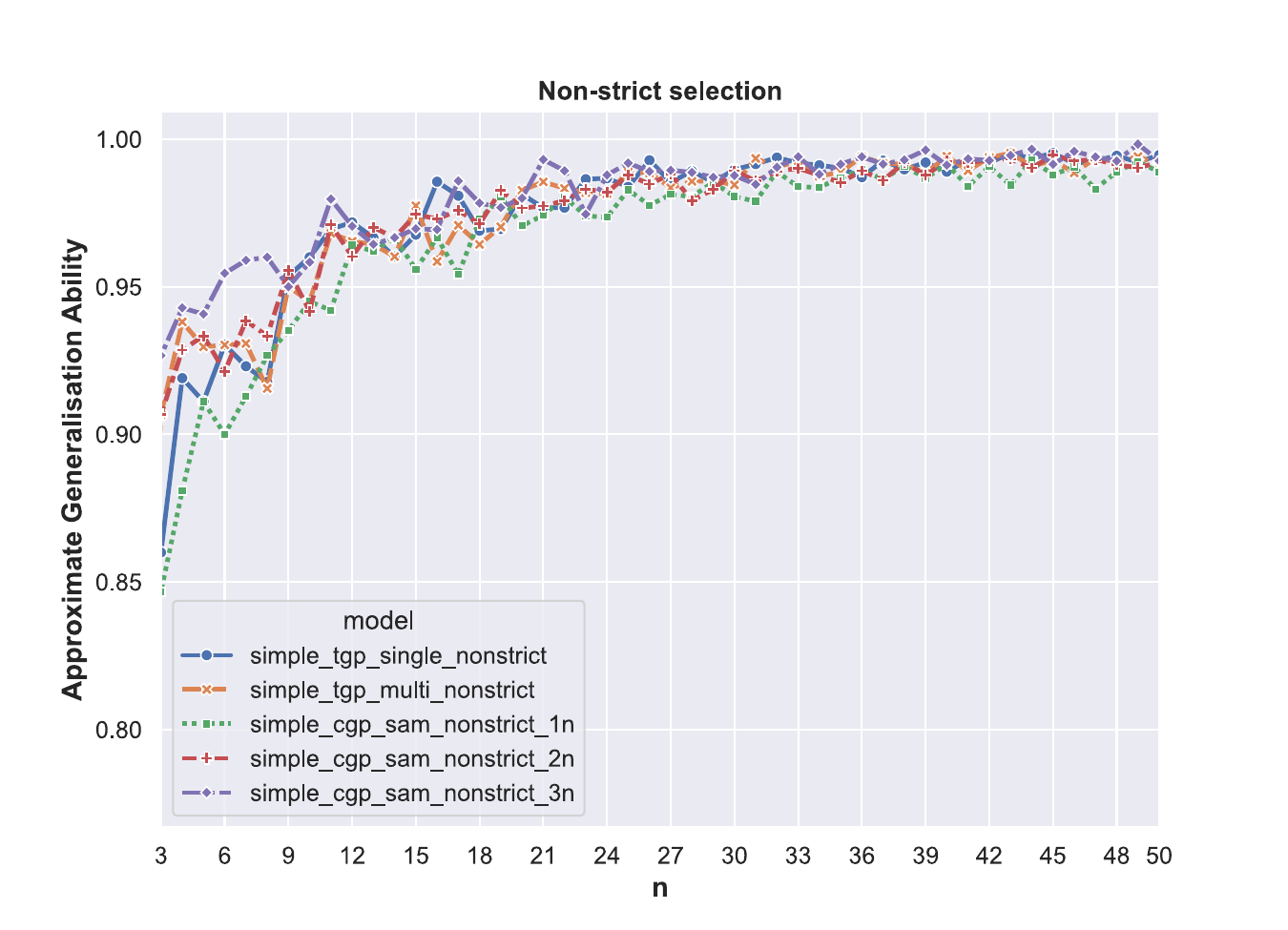} 
\caption{Results of the comparison on \andproblem using incomplete datasets.}
\label{fig:results-runtime-incomplete}
\end{figure}


We evaluated various settings of $n$ in each scenario. For the scenario where the complete training set is used, we evaluated $n \in\{3,\ldots,15\}$. However, since the evaluation of an incomplete training set is less expensive, we evaluated the performance for larger settings of $n \in\{3,\ldots,50\}$ in this scenario.
For each value of $n$, 30 runs were performed with different random seeds per model (CGP and TGP).
The mean of each sample per $n$ was then used for the illustration of our results. 
The GP configuration used in our experiments is summarised in Table~\ref{tab:gp_configuration}.
Running time was measured as the number of iterations for the algorithm 
to fit the training set. 
In our plots, we refer to the TGP models as \texttt{simple\_tgp\_single} and \texttt{simple\_tgp\_multi} and to the CGP model as \texttt{simple\_cgp\_sam}. 
For CGP, we defined the length of the genotype $\Nfunc$ with respect to the dimension $n$ but also considered extra nodes to 
see if there is any gap in runtime. More precisely, we used
$\Nfunc\in\{n, 2n, 3n\}$, and 
refer to as \texttt{1n}, \texttt{2n}, \texttt{3n} in 
our results.


We 
set 
the size of the 
incomplete training and testing sets \newedit{$S$ and $S'$} as $|S| = |S'| = \lceil n^k \rceil$ with $k = 1.3$ 
for 
the settings of large $n$. 
Both
\newedit{sets}
were generated uniformly at random.  
\newedit{Approximating the generalisation ability $G_g(y)$ by $\hat{G}(y,S')$ was done with the formula as explained in Section~\ref{sec:logic-synth}.}
Statistics are reported per setting of $n$ excluding unsuccessful runs, \ie exceeding the maximum number of iterations without fitting the training set. 

\textbf{Results}:  Figure~\ref{fig:results-runtime-complete} shows the results for the running time comparison on \andproblem when the complete training set was used. It can be clearly seen that the considered TGP models perform generally better in this scenario. It can also be seen that increasing the number of function nodes generally improves the search performance of CGP while using strict selection deteriorates its performance. 
Figure~\ref{fig:results-runtime-incomplete} show the results of our experiments with incomplete training and testing sets and it can also be seen that increasing the number of function nodes improves the search performance of CGP in terms of running time for both selection strategies. It is also evident that the generalisation error is relatively low for all models considered demonstrating good generalisation \newedit{abilities} on \andproblem. 
Detailed statistical evaluation of the results can be found in the appendix document. 





\section{Conclusions \newedit{and Discussions}} 

We presented the first runtime analysis of CGP for evolving conjunctions and disjunctions, considering both complete and incomplete training sets. For \andproblem with $n$ variables and $D$ nodes, we proved expected runtimes of $O(nD^5)$ for \opocgpstrict and $O(nD^4)$ for \opocgp with the complete training set.
\newedit{Our analysis revealed the characteristics of CGP induced search that enabling the acceptance of equally good solutions with active gates non-contributing to fitness is beneficial.} 
We also showed that \opocgp fails with overwhelming probability on the complete training set of the \xorproblem.  
To complement our theoretical analysis, we conducted an empirical study examining the runtime performance and generalisation ability of CGP and TGP models for \andproblem. 

\newedit{The theoretical analysis of incomplete training sets for \xorproblem is left out for the reason that the generalisation ability one can hope for is only as good as a random guess. For \andproblem, there is a technical obstacle that on an incomplete training set, our approach using the fitness-level method and Markov chain analysis no longer works with the current level partition. In particular, it is possible to swap a larger set of input variables for a smaller one, 
and such a change still improves the fitness. 
The same issue must occur for TGP under a global mutation operator, however we could not find any literature on this. 
We also believe drift analysis~\cite{DoerrJW2010} is a better-suited tool for addressing such a setting, and leave the analysis for future work.} 
Future work 
\newedit{should also} 
include extending the analysis to general Boolean formulas with AND/OR, identifying broader function classes where CGP is effective, and studying its behaviour on flat fitness landscapes. 
Achieving th\newedit{ese}, however, requires a deeper understanding of the underlying dynamics of CGP, which still is largely unexplored.

\bibliographystyle{splncs04}
\bibliography{references} 

\appendix

\ifarxiv
\section{Useful Tools for the Analysis}

The following tail bounds for sums of geometric random variables are from~\cite{Witt2014}. 
\begin{lemma}[Theorem~1 in~\cite{Witt2014}]\label{lem:tail-bound-geometric}
Let $(X_i)_{i\in[n]}$ be $n$ independent random variables following geometric
distributions with the corresponding success probabilities $(p_i)_{i\in[n]}$,
and define $X:=\sum_{i=1}^{n} X_i$, $p_{\min}:=\min\{p_i\mid i \in[n]\}$.
If $s\geq \sum_{i=1}^{n} (1/p_i^2)$ for some $s<\infty$ then
\[
\prob{X\geq \expect{X}+\gamma} \leq e^{-\frac{\gamma}{4}\cdot \min\left\{\frac{\gamma}{s},p_{\min}\right\}}
\text{ and }
\prob{X\leq \expect{X}-\gamma} \leq e^{-\frac{\gamma^2}{2s}}.
\]
\end{lemma}


The negative drift theorem is from~\cite{OlivetoW2011,OlivetoW2012}.
\begin{theorem}[Theorem~2 in~\cite{OlivetoW2012}]\label{thm:negative-drift}
Let $(X_t)_{t\in \N}$ be a real-value stochastic process over some state space. 
Suppose there exist an interval $[a,b]\subset \R$, two constants $\delta,\varepsilon>0$, and possibly depending on $\ell:=b-a$ a function $r(\ell)$ such that $1\leq r(\ell)=o(\ell/\log{\ell})$ such that for all $t\geq 0$ the following conditions hold:
\begin{enumerate}
\item[1.] $\expect{X_{t+1}-X_{t}\mid X_t, a<X_t<b} \geq \varepsilon$,
\item[2.] $\prob{|X_{t+1}-X_t|\geq j\mid X_t,a<X_t} \leq \frac{r(\ell)}{(1+\delta)^j}$ for $j\in \N$.
\end{enumerate}
Then there exists a constant $c>0$ such that for $T:=\min\{t\geq 0\mid X_t\leq a\}$ it holds 
that $\prob{T\leq 2^{c\ell/r(\ell)}\mid X_0\geq b} = 2^{-\Omega(\ell/r(\ell))}$. 
\end{theorem}

\section{Additional Statistical Evaluation of the Experimental Results}

Table~\ref{tab:results_comparison-complete-runtime} shows the statistical evaluation for the comparison of running time between TGP and CGP on \andproblem using non-strict and strict selection as well as complete datasets. 


\begin{table}[ht]
\centering
\scalebox{\ifarxiv0.44\else0.36\fi}{
    \begin{tabular}{@{\extracolsep{4pt}}lllcccccccc}
\toprule   
 \textbf{Selection} & {$n$} & \textbf{Model} & \textbf{Successful Runs} & \textbf{Avg. Iterations} & \textbf{SD} & \textbf{Min} & \textbf{Q1} & \textbf{Median} & \textbf{Q3} & \textbf{Max} \\ 
\midrule 

\multirow{70}{1in}{\textbf{strict}} & \multirow{5}{1in}{3} & simple\_cgp\_sam\_strict\_1n&30&19&32&0&3&10&17&167 \\ 
& &simple\_cgp\_sam\_strict\_2n&30&8&14&0&0&2&8&61 \\ 
& &simple\_cgp\_sam\_strict\_3n&30&3&5&0&0&0&4&18 \\ 
& &simple\_tgp\_multi\_strict&30&11&12&2&4&8&14&60 \\ 
& &simple\_tgp\_single\_strict&30&12&8&2&6&8&18&30 \\ 

\cline{2-11} 

& \multirow{5}{1in}{4} & simple\_cgp\_sam\_strict\_1n&29&206&483&2&11&29&85&1985 \\ 
& & simple\_cgp\_sam\_strict\_2n&30&174&807&0&2&12&25&4439 \\ 
& & simple\_cgp\_sam\_strict\_3n&30&24&57&0&1&6&14&304 \\ 
& & simple\_tgp\_multi\_strict&30&16&10&1&8&15&22&47 \\ 
& & simple\_tgp\_single\_strict&30&20&10&7&12&18&25&47 \\ 

\cline{2-11} 

& \multirow{5}{1in}{5} &  simple\_cgp\_sam\_strict\_1n&29&2002&7359&1&35&198&833&39909 \\
& &simple\_cgp\_sam\_strict\_2n&30&102&238&4&12&18&50&1062 \\
& &simple\_cgp\_sam\_strict\_3n&30&48&109&0&10&20&33&595 \\
& &simple\_tgp\_multi\_strict&30&22&14&5&8&20&33&54 \\
& &simple\_tgp\_single\_strict&30&29&18&5&16&24&38&80 \\

\cline{2-11}  

& \multirow{5}{1in}{6} &simple\_cgp\_sam\_strict\_1n&20&5055&14315&13&38&224&2418&63331 \\
& &simple\_cgp\_sam\_strict\_2n&30&146&287&2&21&50&103&1396 \\
& &simple\_cgp\_sam\_strict\_3n&30&124&300&0&12&31&96&1322 \\
& &simple\_tgp\_multi\_strict&30&25&15&8&17&24&28&83 \\
& &simple\_tgp\_single\_strict&30&47&32&8&27&42&58&168 \\

\cline{2-11} 

& \multirow{5}{1in}{7} &simple\_cgp\_sam\_strict\_1n&15&34906&80917&33&304&7568&22293&314957 \\
& &simple\_cgp\_sam\_strict\_2n&30&4061&11429&10&83&174&1054&51464 \\
& &simple\_cgp\_sam\_strict\_3n&30&93&132&20&30&58&97&715 \\
& &simple\_tgp\_multi\_strict&30&36&17&10&24&28&50&77 \\
& &simple\_tgp\_single\_strict&30&45&21&16&29&44&59&97 \\

\cline{2-11} 

& \multirow{5}{1in}{8} &simple\_cgp\_sam\_strict\_1n&17&58849&138929&134&344&1137&5741&548412 \\
& &simple\_cgp\_sam\_strict\_2n&30&11533&45056&43&84&436&2148&245973 \\
& &simple\_cgp\_sam\_strict\_3n&30&1122&2840&22&76&144&456&13817 \\
& &simple\_tgp\_multi\_strict&30&40&19&11&24&36&53&91 \\
& &simple\_tgp\_single\_strict&30&66&29&24&43&64&76&149 \\

\cline{2-11} 

& \multirow{5}{1in}{9} 
& simple\_cgp\_sam\_strict\_1n & 0 & -& - & - & - & - & - & - \\
& &simple\_cgp\_sam\_strict\_2n&30&9890&29042&34&96&627&6977&158233 \\
& &simple\_cgp\_sam\_strict\_3n&30&768&2076&13&49&104&218&8521 \\
& &simple\_tgp\_multi\_strict&30&47&16&17&40&47&56&87 \\
& &simple\_tgp\_single\_strict&30&69&30&36&49&57&78&166 \\

\cline{2-11} 

& \multirow{5}{1in}{10}& simple\_cgp\_sam\_strict\_1n & 0 & -& - & - & - & - & - & - \\ 
& & simple\_cgp\_sam\_strict\_2n&30&7539&20298&19&120&517&1982&88716 \\
& &simple\_cgp\_sam\_strict\_3n&30&2561&5308&40&112&306&1645&22711 \\
& &simple\_tgp\_multi\_strict&30&59&25&27&38&56&68&121 \\
& &simple\_tgp\_single\_strict&30&83&37&21&57&71&105&165 \\

\cline{2-11} 

& \multirow{5}{1in}{11} & simple\_cgp\_sam\_strict\_1n & 0 & -& - & - & - & - & - & - \\ 
& &simple\_cgp\_sam\_strict\_2n&30&21380&45778&121&544&6241&15308&232758 \\
& &simple\_cgp\_sam\_strict\_3n&30&15031&50296&26&128&272&3407&271033 \\
& &simple\_tgp\_multi\_strict&30&78&39&24&54&70&91&210 \\
& &simple\_tgp\_single\_strict&30&87&27&31&72&84&103&144 \\

\cline{2-11} 

& \multirow{5}{1in}{12} & simple\_cgp\_sam\_strict\_1n & 0 & -& - & - & - & - & - & - \\ 
& &simple\_cgp\_sam\_strict\_2n&28&29393&38093&72&707&13242&40215&136393 \\
& &simple\_cgp\_sam\_strict\_3n&30&6828&16539&50&292&481&3152&77503 \\
& &simple\_tgp\_multi\_strict&30&81&25&36&62&80&98&129 \\
& &simple\_tgp\_single\_strict&30&122&52&61&80&114&144&263 \\

\cline{2-11} 

&\multirow{5}{1in}{13} & simple\_cgp\_sam\_strict\_1n & 0 & -& - & - & - & - & - & - \\ & &simple\_cgp\_sam\_strict\_2n&30&48360&52182&145&10518&36008&57160&198080 \\
& &simple\_cgp\_sam\_strict\_3n&30&14865&38763&75&350&1405&5457&175512 \\
& &simple\_tgp\_multi\_strict&30&81&32&22&59&79&97&161 \\
& &simple\_tgp\_single\_strict&30&111&50&43&78&101&136&222 \\

\cline{2-11} 

& \multirow{5}{1in}{14} & simple\_cgp\_sam\_strict\_1n & 0 & -& - & - & - & - & - & - \\ 
& &simple\_cgp\_sam\_strict\_2n&22&34233&34499&273&3663&23192&52189&105703 \\
& &simple\_cgp\_sam\_strict\_3n&29&12313&18914&170&649&1603&18752&64404 \\
& &simple\_tgp\_multi\_strict&30&100&34&49&78&94&116&189 \\
& &simple\_tgp\_single\_strict&30&137&37&88&112&135&158&225 \\

\cline{2-11} 

& \multirow{5}{1in}{15} & simple\_cgp\_sam\_strict\_1n & 0 & -& - & - & - & - & - & - \\ 
& &simple\_cgp\_sam\_strict\_2n&18&9820&17151&424&1338&2954&5126&55660 \\
& &simple\_cgp\_sam\_strict\_3n&27&8812&14289&131&406&1103&10978&43166 \\
& &simple\_tgp\_multi\_strict&30&113&43&51&87&108&128&214 \\
& &simple\_tgp\_single\_strict&30&137&56&50&85&133&168&289 \\

\midrule

 \multirow{70}{1in}{\textbf{non-strict}} & \multirow{5}{1in}{3} &
simple\_cgp\_sam\_nonstrict\_1n&30&11&15&0&1&7&14&52 \\
& & simple\_cgp\_sam\_nonstrict\_2n&30&3&5&0&0&2&5&23 \\
& & simple\_cgp\_sam\_nonstrict\_3n&30&4&6&0&0&1&4&29 \\
& & simple\_tgp\_multi\_nonstrict&30&10&7&1&5&8&14&31 \\
& & simple\_tgp\_single\_nonstrict&30&14&9&1&6&12&22&42 \\

\cline{2-11} 

& \multirow{5}{1in}{4} & simple\_cgp\_sam\_nonstrict\_1n&30&43&41&2&18&29&52&200 \\
& &simple\_cgp\_sam\_nonstrict\_2n&30&17&21&0&4&9&20&75 \\
& &simple\_cgp\_sam\_nonstrict\_3n&30&15&23&0&2&6&17&101 \\
& &simple\_tgp\_multi\_nonstrict&30&14&8&2&7&12&18&32 \\
& &simple\_tgp\_single\_nonstrict&30&22&16&4&11&16&28&71 \\

\cline{2-11} 

&  \multirow{5}{1in}{5} &simple\_cgp\_sam\_nonstrict\_1n&30&128&169&2&33&68&109&710 \\
& &simple\_cgp\_sam\_nonstrict\_2n&30&62&103&0&19&25&53&493 \\
& &simple\_cgp\_sam\_nonstrict\_3n&30&26&29&0&10&17&34&148 \\
& &simple\_tgp\_multi\_nonstrict&30&21&10&8&13&18&29&42 \\
& &simple\_tgp\_single\_nonstrict&30&28&14&7&19&24&37&58 \\

\cline{2-11} 

& \multirow{5}{1in}{6} &simple\_cgp\_sam\_nonstrict\_1n&30&213&179&25&72&131&325&717 \\
& &simple\_cgp\_sam\_nonstrict\_2n&30&66&45&0&39&58&86&181 \\
& &simple\_cgp\_sam\_nonstrict\_3n&30&45&36&10&20&35&61&157 \\
& & simple\_tgp\_multi\_nonstrict&30&27&15&5&15&24&40&60 \\
& &simple\_tgp\_single\_nonstrict&30&34&16&11&22&32&46&84 \\

\cline{2-11} 

& \multirow{5}{1in}{7} & simple\_cgp\_sam\_nonstrict\_1n&30&319&303&26&134&188&438&1436 \\
& &simple\_cgp\_sam\_nonstrict\_2n&30&115&98&6&45&98&138&386 \\
& &simple\_cgp\_sam\_nonstrict\_3n&30&64&61&1&28&46&71&281 \\
& &simple\_tgp\_multi\_nonstrict&30&35&16&6&27&32&48&74 \\
& &simple\_tgp\_single\_nonstrict&30&54&32&26&33&44&66&187 \\

\cline{2-11} 

& \multirow{5}{1in}{8}&simple\_cgp\_sam\_nonstrict\_1n&30&623&732&56&254&396&696&3812 \\
& &simple\_cgp\_sam\_nonstrict\_2n&30&150&127&5&57&104&226&525 \\
& &simple\_cgp\_sam\_nonstrict\_3n&30&118&135&18&34&70&130&608 \\
& &simple\_tgp\_multi\_nonstrict&30&38&16&17&25&38&50&67 \\
& &simple\_tgp\_single\_nonstrict&30&67&33&17&46&60&84&164 \\

\cline{2-11} 

& \multirow{5}{1in}{9} &simple\_cgp\_sam\_nonstrict\_1n&30&1248&1046&95&294&1124&2025&3372 \\
& &simple\_cgp\_sam\_nonstrict\_2n&30&296&225&43&118&196&444&967 \\
& &simple\_cgp\_sam\_nonstrict\_3n&30&139&110&22&64&104&192&524 \\
& &simple\_tgp\_multi\_nonstrict&30&49&27&15&33&44&58&166 \\
& &simple\_tgp\_single\_nonstrict&30&79&36&32&52&70&94&191 \\

\cline{2-11} 

& \multirow{5}{1in}{10} &simple\_cgp\_sam\_nonstrict\_1n&30&1419&1042&152&701&1311&1679&3986 \\
& &simple\_cgp\_sam\_nonstrict\_2n&30&326&234&41&153&264&404&1015 \\
& &simple\_cgp\_sam\_nonstrict\_3n&30&295&399&49&110&161&256&1792 \\
& &simple\_tgp\_multi\_nonstrict&30&52&19&16&38&48&63&101 \\
& &simple\_tgp\_single\_nonstrict&30&86&33&32&64&80&99&172 \\

\cline{2-11} 

& \multirow{5}{1in}{11} &simple\_cgp\_sam\_nonstrict\_1n&30&2348&1880&329&1081&1970&2909&9072 \\
& &simple\_cgp\_sam\_nonstrict\_2n&30&375&281&68&191&298&502&1101 \\
& &simple\_cgp\_sam\_nonstrict\_3n&30&234&199&26&86&166&302&781 \\
& &simple\_tgp\_multi\_nonstrict&30&57&22&20&40&55&70&128 \\ 
& &simple\_tgp\_single\_nonstrict&30&108&46&47&72&94&134&218 \\

\cline{2-11} 

& \multirow{5}{1in}{12}&simple\_cgp\_sam\_nonstrict\_1n&30&2824&3513&240&1185&1772&2374&18617 \\
& &simple\_cgp\_sam\_nonstrict\_2n&30&595&358&60&344&528&790&1545 \\
& &simple\_cgp\_sam\_nonstrict\_3n&30&317&233&35&127&242&465&806 \\
& &simple\_tgp\_multi\_nonstrict&30&69&26&28&52&67&81&126 \\
& &simple\_tgp\_single\_nonstrict&30&112&41&51&80&100&141&206 \\

\cline{2-11} 

& \multirow{5}{1in}{13} &simple\_cgp\_sam\_nonstrict\_1n&30&2732&1854&440&1395&2344&3506&8635 \\
& &simple\_cgp\_sam\_nonstrict\_2n&30&969&1097&44&272&714&1345&5627 \\
& &simple\_cgp\_sam\_nonstrict\_3n&30&424&373&38&158&298&472&1402 \\
& &simple\_tgp\_multi\_nonstrict&30&71&28&31&53&63&87&127 \\
& &simple\_tgp\_single\_nonstrict&30&120&40&45&91&108&141&241 \\

\cline{2-11} 

& \multirow{5}{1in}{14} &simple\_cgp\_sam\_nonstrict\_1n&30&4608&3438&767&2576&3742&6440&15033 \\
& &simple\_cgp\_sam\_nonstrict\_2n&30&807&738&140&396&612&1092&3992 \\
& &simple\_cgp\_sam\_nonstrict\_3n&30&504&662&55&196&338&457&3030 \\
& &simple\_tgp\_multi\_nonstrict&30&86&30&39&65&80&101&167 \\
& &simple\_tgp\_single\_nonstrict&30&122&39&56&95&122&148&226 \\

\cline{2-11} 

& \multirow{5}{1in}{15}&simple\_cgp\_sam\_nonstrict\_1n&30&4782&3219&698&2870&3598&5304&13031 \\
& &simple\_cgp\_sam\_nonstrict\_2n&30&1093&999&229&532&754&1206&5107 \\
& &simple\_cgp\_sam\_nonstrict\_3n&30&640&504&111&277&531&815&2240 \\
& &simple\_tgp\_multi\_nonstrict&30&96&21&39&84&96&112&144 \\
& &simple\_tgp\_single\_nonstrict&30&135&43&68&110&121&159&219 \\

\bottomrule
\end{tabular}

}
\caption{Results for the runtime comparison between TGP and CGP on \andproblem for $n \in \{3, \ldots , 15\}$ using complete datasets. Note that for \texttt{1n}, CGP 
failed to evolve solutions for $n > 9$ with strict selection.} 
\label{tab:results_comparison-complete-runtime}
\end{table}
\fi

\end{document}